# Decision Tree and K-Means Analysis of Raman Spectra for Edible Oils: A Physics-Informed AI Approach

Amrita Shaw[a], Chandrasekar S. N.[b], Sai Muthukumar V.[c*], Jhinuk Gupta[a*,] Deepak L. N. Kallepalli[d*]

[a] Department of Food and Nutritional Sciences, Sri Sathya Sai Institute of Higher Learning, Vidyagiri, Prasanthi Nilayam, Sri Sathya Sai District, Andhra Pradesh, India – 515134.

[b]Aix-Marseille University, Marseille, 13007, France.

[c]Department of Physics, Sri Sathya Sai Institute of Higher Learning, Vidyagiri, Prasanthi Nilayam, Sri Sathya Sai District, Andhra Pradesh, India – 515134.

[d]CogniEvolve AI Inc., 410-25 Woodridge Cres., Nepean K2B 7T4, ON, Canada.

*Corresponding author(s) email ID: klndphyai@gmail.com, vsaimuthukumar@sssihl.edu.in, jhinukgupta@sssihl.edu.in,

## Abstract

Classification of edible oils in processed foods is important for food quality, fraud prevention, and regulatory compliance. This study establishes a MECE (Mutually Exclusive, Collectively Exhaustive) analytical framework that systematically integrates intrinsic spectral organization, interpretable classification, Physics-Informed Artificial Intelligence (PI-AI), and Frugal AI-based feature reduction for edible-oil classification. Five edible oils were investigated in pure form and within a fried-potato-chip matrix using t-SNE, K-means clustering, Decision Trees, and Non-Negative Least Squares (NNLS)-based spectral decomposition. Unsupervised analyses revealed substantially stronger class organization and separability in pure oils, whereas food-matrix effects introduced pronounced spectral overlap. Decision Trees achieved 100% classification accuracy for pure oils using only four Raman variables from the original 1866-feature spectral space. These four variables, consistently identified by both pre-pruned and post-pruned models, represented only approximately 0.21% of the available spectral information while retaining perfect test-set performance. Notably, two Raman variables associated with lipid unsaturation (~1650 $cm^{-1}$) and hydrocarbon-chain skeletal organization (~1127 $cm^{-1}$) remained important after NNLS-based matrix correction, with their combined contribution increasing from 50% in pure oils to approximately 62% and 89% in the paper-subtracted and paper-potato-subtracted datasets, respectively, indicating the emergence of robust oil-specific signatures in complex food matrixes. For the food matrix samples, NNLS-based PI-AI spectral decomposition substantially improved classification by separating oil-related signatures from paper and potato contributions. Optimized post-pruned models achieved test accuracies of nearly 80% for paper-subtracted and paper-plus-potato-subtracted datasets, while reducing the number of important Raman variables to only five and four. The compact four-feature representation further reduced the data footprint by 99.44% without loss of classification accuracy. Collectively, these findings demonstrate that accurate Raman-based oil identification can be achieved through physically meaningful, highly compact, and interpretable spectral representations, providing a promising foundation for Frugal AI, Edge AI, portable sensing, and embedded food-quality monitoring.

## 1. Introduction

The identification of edible oils in food products is important for ensuring product quality, consumer safety, and regulatory compliance. Differences in fatty-acid composition, degree of unsaturation, and processing history influence both nutritional value and physicochemical properties, while adulteration, oil substitution, and repeated frying practices can compromise food quality and consumer trust [1–6]. Conventional analytical techniques such as gas chromatography (GC), high-performance liquid chromatography (HPLC), and nuclear magnetic resonance (NMR) provide reliable compositional information but generally require extensive sample preparation, solvent extraction, and laboratory-based analysis [4, 5]. Raman spectroscopy has emerged as a promising alternative because it is rapid, non-destructive, and capable of providing molecular-level information related to lipid composition through characteristic vibrational bands associated with C=C stretching, $CH_2$ deformation, and ester functional groups [6–8]. Consequently, Raman spectroscopy has gained increasing attention for edible-oil classification and food-quality assessment.

To exploit the rich spectral information generated by Raman measurements, a wide range of chemometric and machine-learning approaches have been applied. Principal Component Analysis (PCA), Analysis of Variance (ANOVA), Multivariate Analysis of Variance (MANOVA), and related chemometric methods have been widely used to investigate spectral variability and establish statistically significant class separation [9–11]. Likewise, supervised learning algorithms including Support Vector Machines (SVM), Random Forests, Decision Trees, and ensemble learning methods have demonstrated strong predictive capabilities for edible-oil classification based on spectroscopic data [12–15]. Although many studies report high classification accuracy for pure oils, the performance of these approaches often declines when applied to real food systems because spectral contributions from starch, proteins, moisture, seasonings, and other matrix components increase class overlap and reduce discrimination power [7,13,16].

Our previous studies contributed to this field from two complementary perspectives. In the first study, a Raman–chemometric framework based on chemically interpretable ratio-metric spectral markers was developed to differentiate five edible oils in both pure and food-matrix samples using a rapid, solvent-free sampling strategy [1]. The study demonstrated that carefully selected Raman marker ratios combined with PCA and MANOVA could provide statistically significant differentiation while maintaining strong chemical interpretability [1]. Subsequently, we extended this work using machine-learning approaches and demonstrated that supervised classification models, particularly kernel-based algorithms, could accurately identify edible oils from both pure-oil and fried-food datasets, thereby confirming that oil-specific Raman signatures remain detectable even within complex food matrices [2]. Together, these studies established the chemical basis and predictive capability of Raman-based edible-oil classification while highlighting the influence of food-matrix effects on classification performance.

While these studies successfully established both chemically interpretable Raman markers and highly accurate machine-learning classifiers, they did not explicitly examine how the intrinsic organization of the Raman spectral space influences downstream classification performance. Consequently, an important question remains unresolved: whether predictive success arises primarily from the choice of learning algorithm or from the degree of natural class structure present within the spectral data itself. Addressing this question is important for developing machine-learning models that are not only accurate but also interpretable and transferable across different food matrices.

Despite these advances, the relationship between the intrinsic spectral structure of Raman data and the resulting classification performance remains insufficiently understood. Most studies, including our previous work, focus primarily on statistical separation or predictive accuracy without first examining whether the underlying spectral data exhibit natural class organization. This distinction is particularly important because the success of supervised learning fundamentally depends on the degree of separability present within the feature space [17–19]. Therefore, the present study introduces a unified analytical framework that integrates unsupervised and supervised learning for Raman-based edible-oil identification. K-means clustering and t-distributed stochastic neighbor embedding (t-SNE) are first employed to evaluate the natural grouping behavior and separability of pure-oil and fried-food spectral datasets, followed by Decision Tree classification to generate transparent and interpretable predictive models.

From a methodological perspective, the proposed framework follows a Mutually Exclusive, Collectively Exhaustive (MECE) design philosophy, in which each analytical stage performs a distinct and non-overlapping role while collectively covering the complete pathway from intrinsic spectral organization to interpretable classification [20]. Specifically, t-SNE and K-means assess natural spectral structure, NNLS-based Physics-Informed AI attenuates spectral variation associated with the food-matrix components (paper and potato), and Decision Trees evaluate predictive separability and feature-level interpretability. Together, these stages provide a comprehensive understanding of how spectral organization influences machine-learning performance. This is related to the use of complete orthogonal basis sets in quantum mechanics and complete phase-space descriptions in statistical physics, where non-overlapping components collectively provide a full representation of the system [21-22].

Recent developments in Physics-Informed Artificial Intelligence (PI-AI) have demonstrated that incorporating established scientific principles into machine-learning workflows can improve model robustness, generalizability, and interpretability [23-24]. Unlike purely data-centric machine-learning approaches that rely exclusively on statistical relationships present in training datasets, PI-AI integrates prior domain knowledge, physical constraints, conservation laws, governing equations, or mechanistic understanding directly into the learning process. By restricting the solution space to physically plausible outcomes, PI-AI can reduce overfitting, improve performance when data are limited or noisy, and generate models whose predictions remain consistent with known scientific behaviour. The growing importance of this paradigm has been widely recognized across scientific machine learning, where physics-informed approaches have been shown to effectively bridge the gap between empirical observations and mechanistic understanding.

In Raman spectroscopy, the measured spectrum can be viewed as an additive combination of contributions from multiple chemical constituents, making non-negative spectral decomposition a physically meaningful modeling strategy. Consequently, Non-Negative Least Squares (NNLS) provides a natural PI-AI framework because it preserves the additive nature of spectral mixing while enforcing the physically realistic constraint that component contributions cannot be negative. The standard NNLS algorithm was introduced by Lawson and Hanson, while Bro and De Jong subsequently developed a faster implementation, Fast Non-Negative Least Squares (FNNLS), for chemometric applications. FNNLS has since become a widely adopted approach for extracting chemically interpretable constituent contributions from complex spectroscopic mixtures [25]. In the present study, NNLS was employed as a physics-informed feature extraction strategy to reduce food-matrix interference by separating paper and potato spectral contributions from the oil-related Raman signal. The resulting chemically meaningful difference spectra were subsequently analyzed using unsupervised clustering and interpretable Decision Tree models. By explicitly linking intrinsic spectral organization, physics-informed spectral decomposition, and supervised classification performance, this work provides new insight into how spectral structure governs machine-learning success and advances the development of robust, interpretable approaches for edible-oil classification in both ideal and complex food systems.

Recent interest in Frugal AI has emphasized the development of machine-learning systems that achieve reliable performance while minimizing data requirements, model complexity, computational cost, and energy consumption. Rather than pursuing increasingly large and resource-intensive models, Frugal AI seeks compact and efficient representations that remain accurate, interpretable, and suitable for deployment in resource-constrained environments [26-27]. This idea aligns naturally with spectroscopic analytics, where identification of a small number of highly informative spectral variables can enable fast and computationally efficient classification without sacrificing predictive performance. Beyond edge and embedded sensing, Frugal AI has also been identified as a key enabler of Industry 4.0, where lightweight, low-power models are needed to process the large volumes of real-time sensor data generated in manufacturing and food-processing environments; compact classifiers such as the one developed here are therefore well suited to in-line quality-control and classification tasks within food-processing lines [26].

## 2. Materials and Methods

### 2.1 Sample Collection and Preparation

Five commercially available edible oils were investigated in this study: sunflower oil (SO), soybean oil (SOYO), groundnut oil (GNO), palm oil (PO), and vanaspati oil (VO). Spectral datasets were acquired for both pure oil samples and oils extracted from a representative fried-food matrix. The sampling strategy, oil selection criteria, and Raman measurement protocol followed the methodologies established in our previous studies, which demonstrated the feasibility of solvent-free Raman analysis for edible-oil classification in both pure and food-matrix conditions [1,2].

To evaluate the influence of matrix complexity on spectral separability and classification performance, four datasets were constructed (CSV files in ref. [28]): (i) a pure-oil dataset representing ideal measurement conditions, (ii) a fried-food dataset representing a realistic food-classification scenario, (iii) NNLS-corrected dataset for fried-food dataset by removing the paper contribution, and (iv) NNLS-corrected dataset for fried-food dataset by removing the paper and starch (from potato) contribution. The details of data preparation are explained in [2]. The pure-oil dataset consisted of 1,000 Raman spectra, with 200 spectra collected for each of the five oil classes, whereas the fried-food dataset consisted of 900 Raman spectra, with 180 spectra collected for each oil class (complete details are in [1]). This balanced experimental design enabled direct comparison of class separability and classification performance between pure-oil and food-matrix conditions.

### 2.2 Raman Spectral Acquisition

Raman spectra were acquired using the instrumentation and acquisition parameters described in our previous studies [1, 2]. Measurements were performed using a rapid, solvent-free sampling approach requiring minimal sample preparation. Spectra were collected over the fingerprint and high-wavenumber regions, covering approximately 500–3000 $cm^{-1}$, thereby capturing vibrational bands associated with C=C stretching, $CH_2$ deformation, and other molecular vibrations characteristic of edible oils. The collected spectra were stored as intensity values corresponding to individual Raman shifts and subsequently organized into a feature matrix for computational analysis. Each spectrum was assigned a class label corresponding to its oil type and sample category (pure oil or fried-food matrix).

### 2.3 Spectral Preprocessing

Prior to machine-learning analysis, spectral preprocessing was performed to improve comparability across samples. This data (each spectrum) is already baseline-corrected, smoothed, and Z-score normalized where each wavenumber/feature column is centered and scaled across the dataset. **Figure 1** shows the mean Raman spectra plots generated for each oil class. These processed spectra served as the input for all clustering and classification analyses presented in this study. The spectral region from 1900 to 2600 $cm^{-1}$ was not shown in the **Figure 1** as it does not contain any useful information.

## 2.4 Physics-Informed Spectral Decomposition Using NNLS

To reduce matrix-induced spectral variability, a physics-informed feature extraction strategy (Physics-Informed AI) based on Non-Negative Least Squares (NNLS) was implemented. NNLS models an observed Raman spectrum as a linear combination of reference spectral components while enforcing non-negative coefficients:

$$X_{chips} = C_{oil} (X_{oil}) + C_{potato} (X_{potato}) + C_{paper} (X_{paper}) + \varepsilon \text{ – (Equation 1)}$$

subject to:

$$C_{oil}, C_{potato}, C_{paper} \geq 0 \text{ – (Equation 2)}$$

where $X_{chips}$ represents the measured spectrum, $X_{oil,}$ $X_{potato}$, and $X_{paper}$ contain the reference Raman spectra of unheated oil, dehydrated dry potato, and paper. The contribution coefficients $C_{oil}$, $C_{potato}$, and $C_{paper}$ are estimated through constrained least-squares optimization. The non-negativity constraint reflects the physical reality that Raman spectral contributions and chemical concentrations cannot be negative.

In the present work, each fried-food Raman spectrum was decomposed using an oil reference together with fixed paper and potato reference spectra. NNLS assumes that the measured spectrum can be approximated as an additive linear combination of these reference spectra and constrains their fitted coefficients to be non-negative. The fitted paper and potato terms were subsequently subtracted to generate matrix-corrected Raman spectra for downstream clustering and Decision Tree classification. Thus, NNLS served as a physics-informed spectral preprocessing method.

For each fried-chip Raman spectrum, NNLS was used to express the measured signal as an additive, non-negative combination of three reference spectra representing oil, paper, and potato, reflecting the physical constraint that their fitted contributions cannot be negative. Prior to fitting, all spectra were aligned, baseline-corrected using Asymmetric Least Square (ALS), smoothed, normalized, and interpolated onto a common Raman-shift axis to ensure that the decomposition operated on directly comparable signals. For each chip Raman spectrum, the oil reference was selected as the fresh, unheated replicate of the corresponding known oil type showing the highest correlation with that spectrum, while the paper and potato references were fixed and represented fresh paper tissue and dehydrated potato, respectively. Fitting Equation (1) subject to Equation (2) yielded the spectral scaling coefficients $c_{oil}$, $c_{paper}$, and $c_{potato}$.

From this decomposition, two complementary background-subtracted datasets were generated to isolate the oil-related Raman signal from matrix interference. The first, denoted paper-potato-subtracted, was obtained by subtracting both fitted background terms from the original chip spectrum, thereby preferentially retaining the oil-associated signal together with any unexplained residual. The second, denoted paper-subtracted, removed only the paper contribution, leaving both the oil and potato spectral contributions intact. Comparing model performance on these two datasets allows direct assessment of how much oil-related spectral information is recovered from the chip matrix once the confounding paper and/or potato background is removed, and provides a basis for evaluating the robustness of the physics-informed spectral correction described above. Both these datasets were provided on GitHub [28].

## 2.5 Computational Environment and Reproducibility

All data processing, spectral analysis, unsupervised learning, physics-informed NNLS decomposition, and Decision Tree modeling were performed using Python within Jupyter Notebook and Visual Studio Code (VS Code) environments. To ensure complete reproducibility and consistency of results across all analyses, identical software versions and package dependencies were maintained throughout the study. The computational environment consisted of Python 3.13.9 (Anaconda distribution), Pandas 2.3.3, NumPy 2.3.5, Matplotlib 3.10.6, Seaborn 0.13.2, Plotly 6.3.0, and Scikit-learn 1.7.2. The versions of all software libraries were fixed and verified prior to model development to prevent discrepancies arising from package updates or dependency conflicts. The computational environment used in this work is summarized in [28].

In the interest of transparency and reproducible research, all Jupyter notebooks used for data preprocessing, clustering, NNLS-based spectral decomposition, feature extraction, and machine-learning analysis are provided in [28]. In addition, the corresponding Visual Studio Code project files have been included to facilitate future deployment and practical implementation. While Jupyter notebooks provide an interactive environment for scientific analysis and result exploration, Visual Studio Code offers advantages for software engineering workflows, application development, and deployment-oriented implementation. Providing both environments enables other researchers to reproduce the reported results, validate the analyses, and readily adapt the workflow for future Raman spectroscopy, Physics-Informed Artificial Intelligence (PI-AI), or embedded machine-learning applications.

Several Decision Tree hyperparameters were held constant across all datasets and model configurations to ensure consistency of the comparative analysis [28-29]. class_weight = balanced automatically adjusts class weights according to class frequencies; criterion = gini uses Gini impurity to evaluate the quality of candidate splits; max_features = None allows all available features to be considered when searching for the best split; min_impurity_decrease = 0 requires no minimum impurity reduction beyond the standard splitting criterion; min_samples_leaf = 1 permits a terminal node to contain a single sample; min_weight_fraction_leaf = 0 imposes no minimum weighted fraction of samples in a leaf node; monotonic_cst = None specifies no monotonic constraints on the input features; random_state = 0 ensures reproducible model construction; and splitter = best selects the best available split at each node. By keeping these parameters fixed (some by default), the comparison among the baseline, pre-pruned, and post-pruned models primarily reflects the influence of the hyperparameters intentionally optimized for controlling tree complexity [29].

Overall, the hyperparameter strategy followed a progressive model-optimization framework. The initial Decision Tree provided a reference for evaluating the classification problem without additional complexity constraints, pre-pruning-controlled tree growth through restrictions on depth, leaf nodes, and node splitting, and post-pruning further simplified the models through cost-complexity optimization using ccp_alpha. The dataset-specific configurations summarized in **Table S1** therefore provide an important record of the optimization process and demonstrate that different levels of spectral and food-matrix complexity require different levels of Decision Tree regularization. This systematic approach enables a direct comparison between model complexity, classification performance, and the effectiveness of NNLS-based physics-informed matrix correction.

## 3. Results and Discussion

### 3.1 t-SNE Visualization of Pure Oils and Fried-Food Samples

To examine whether the Raman datasets possessed an inherent class structure before clustering or classification, t-distributed stochastic neighbor embedding (t-SNE) was applied to the standardized high-dimensional spectral features [18]. t-SNE is a nonlinear dimensionality-reduction technique that projects high-dimensional data into a low-dimensional space while preserving local neighborhood relationships among samples. As a result, spectra with similar characteristics tend to appear close together in the visualization, making t-SNE particularly useful for exploring class organization and separability in complex datasets. The purpose of this analysis was purely exploratory, namely, to visualize the local organization of the Raman data and assess whether the five oil classes already formed natural groupings in low-dimensional space. The resulting embeddings for the pure-oil and fried-food datasets are shown in **Figure 2**.

The t-SNE map for the pure-oil dataset reveals a comparatively well-organized arrangement of the five oil classes. Sunflower oil (SO) occupies a clearly distinguishable region of the embedded space, while vanaspati oil (VO) also appears in a relatively compact and separated location. Groundnut oil (GNO) forms another coherent group, indicating that these oils retain strong class-specific Raman signatures. In contrast, palm oil (PO) and vanaspati oil (VO) show partial proximity, and less overlap is also visible between some neighboring classes, especially for SOYO, which suggests that chemically related oils can still exhibit spectral similarity despite overall separability.

The fried-food dataset shows a noticeably weaker class structure. Although the five oil categories remain identifiable in the embedding, the class regions are broader, less compact, and more intermingled than in the pure-oil case. The overlap among oil classes is more pronounced, which is consistent with the presence of matrix-derived spectral contributions originating from the food substrate and frying-induced changes in the sample composition. In this setting, the oil-related Raman fingerprints are still present, but they are partially masked by additional variability from the food matrix, reducing the clarity of class boundaries.

A particularly important feature of the t-SNE visualization is that it suggests a clear contrast between the two analytical scenarios. In pure oils, the spectral signatures are strong enough to generate distinct low-dimensional groupings, whereas in fried-food samples the same signatures become less sharply defined because the matrix introduces overlap between classes. This does not imply that the oil identity is lost in the fried-food system; rather, it indicates that the spectral space becomes less separable and therefore more challenging for machine-learning methods to partition. The visual difference between the two datasets therefore provides an early indication that pure oils contain a stronger intrinsic class structure than fried-food samples.

It is also important to note that t-SNE was used only as a visualization tool and not as the basis for clustering. The embeddings shown in **Figure 2** are intended to reveal neighborhood relationships and latent organization in the Raman data, not to serve as direct inputs to the K-means algorithm. Even so, the observed patterns are informative: the pure-oil map suggests stronger natural separability, while the fried-food map shows the effect of matrix complexity on the same underlying oil signatures (default perplexity of 30). These qualitative observations justify the subsequent unsupervised clustering analysis, which evaluates the extent to which the apparent visual organization corresponds to measurable cluster structure in the original spectral space.

## 3.2 Influence of t-SNE Perplexity on Cluster Stability

The interpretation of t-distributed stochastic neighbor embedding (t-SNE) can be influenced by the choice of the perplexity parameter, which controls the effective number of local neighbors considered when constructing the probability distribution of pairwise relationships within the dataset. Introduced by van der Maaten and Hinton, perplexity is commonly viewed as a measure that balances local and global structure in the low-dimensional representation, with lower values emphasizing local neighborhood relationships and higher values incorporating a broader view of the data manifold [18]. Mathematically, perplexity is defined as

$$Perp(P_i) = 2^{H(P_i)} \text{ – (Equation 3)}$$

where $H(P_i)$represents the Shannon entropy of the conditional probability distribution associated with data point $i$:

$$H(P_i) = -\sum_j p_{j|i} \log_2 p_{j|i} \quad \text{– (Equation 4)}$$

Here, $p_{j|i}$denotes the conditional probability that point $i$selects point $j$as its neighbor in the high-dimensional space. Consequently, perplexity may be interpreted as the effective neighborhood size used by t-SNE during manifold construction. Because the choice of perplexity can influence the appearance of low-dimensional embeddings, evaluating multiple perplexity values is often recommended to assess the robustness of observed clustering patterns [18].

To determine whether the grouping observed in **Figure 2** was dependent on a particular parameter selection, the t-SNE analysis was repeated using perplexities of 5, 10, 20, 30, 50, and 75 for both datasets. The resulting embeddings for the pure-oil dataset are presented in **Figure S1**. Across all tested perplexities, the principal organization of the oil classes remained largely preserved. Sunflower oil (SO), groundnut oil (GNO), and vanaspati oil (VO) consistently occupied relatively distinct regions of the embedded space, while palm oil (PO) maintained partial proximity to neighboring classes. Although the exact geometry and spacing of clusters varied slightly with perplexity, the overall class structure remained stable. This consistency suggests that the observed separation of the pure-oil spectra originates from genuine organization within the Raman data rather than from a specific t-SNE parameter choice.

The fried-food dataset exhibited a different behavior, as shown in **Figure S2**. While class-associated regions remained visible across all perplexity values, the embeddings consistently showed broader distributions and greater inter-class overlap than those observed for pure oils. Changes in perplexity altered the visual arrangement of the clusters but did not substantially improve class separation, indicating that the reduced grouping quality is an intrinsic characteristic of the dataset itself rather than an artifact of parameter selection. The persistence of overlapping regions across the entire perplexity range suggests that matrix-derived spectral variability weakens the local neighborhood structure that t-SNE seeks to preserve in the low-dimensional representation.

To further examine the organization of the spectral data, three-dimensional t-SNE embeddings were generated using a perplexity of 50 (**Figure S3**). The additional dimension provided an alternative view of class arrangement and confirmed the trends observed in the two-dimensional projections. The pure-oil dataset continued to exhibit more compact and distinguishable class regions, whereas the fried-food samples remained comparatively diffuse. Thus, both the two-dimensional and three-dimensional visualizations support the conclusion that the Raman spectra of pure oils possess a stronger and more stable underlying class structure than those obtained from the fried-food matrix.

Overall, the perplexity study demonstrates that the principal observations obtained from t-SNE are robust across a broad range of neighborhood assumptions. The stability of the pure-oil embeddings and the persistent overlap observed in the fried-food dataset indicate that the differences between the two systems reflect genuine variations in spectral organization rather than visualization artifacts. This robustness provides confidence that the separability patterns observed in the t-SNE analysis can be meaningfully investigated using quantitative clustering approaches in the subsequent sections.

### 3.3 Elbow Analysis and Silhouette Scores for K-Means Clustering

While t-SNE provides a qualitative visualization of spectral organization, quantitative evaluation of cluster structure requires objective clustering metrics. Therefore, K-means clustering was applied to the standardized Raman datasets to assess whether the observed class organization could be recovered in the original high-dimensional feature space. K-means is an unsupervised learning algorithm that partitions samples into groups (clusters) based on spectral similarity without using the true labels. Samples assigned to the same cluster are expected to be more similar to one another than to samples belonging to different clusters. The number of clusters was initially explored using the elbow method based on the within-cluster sum of squares (WCSS), followed by silhouette analysis to evaluate cluster compactness and separation. K-means clustering partitions a dataset into $k$ clusters by minimizing the within-cluster variance and has remained one of the most widely used unsupervised learning algorithms since its introduction by MacQueen [17].

Because K-means attempts to group similar samples around cluster centers (centroids), clustering quality can be evaluated by measuring how tightly samples are grouped within each cluster and how well different clusters are separated from one another. The WCSS metric measures the total squared distance between each data point and the centroid of the cluster to which it is assigned:

$$WCSS = \sum_{i=1}^{k} \sum_{x \in C_i} || x - \mu_i ||^2 \quad \text{– (Equation 5)}$$

where $C_i$ represents cluster $i$, $\mu_i$ denotes the centroid of cluster $i$, and $k$ is the number of clusters. Lower WCSS values indicate greater cluster compactness. In the elbow method, WCSS is calculated across a range of cluster numbers, and the optimal solution is often identified near the point where additional clusters provide diminishing improvement in variance reduction [16,17].

The elbow plots obtained for the pure-oil and fried-food datasets are shown in **Figure 3 (a)**. In both datasets, WCSS decreased continuously with increasing cluster number, reflecting the expected improvement in cluster compactness as additional centroids are introduced. For the pure-oil dataset, the reduction in WCSS was relatively pronounced up to approximately five clusters, after which the rate of improvement became more gradual. A similar trend was observed for the fried-food dataset, although the transition was less distinct and the curve exhibited a smoother decline. This behavior suggests that the underlying class structure of the pure-oil dataset is more clearly defined, whereas the fried-food dataset contains a more diffuse organization of samples within the spectral feature space. Consistent with the known experimental design, five clusters were selected for subsequent analysis to enable direct comparison between the unsupervised clustering results and the five oil categories investigated in this study.

Although WCSS measures cluster compactness, it does not directly indicate whether neighboring clusters overlap with one another. Therefore, silhouette analysis was additionally employed to evaluate both cluster compactness and cluster separation simultaneously. The silhouette coefficient measures how well each sample fits within its assigned cluster compared with neighboring clusters [30]. The silhouette coefficient for a data point is defined as

$$s(i) = \frac{b(i) - a(i)}{\max[a(i), b(i)]} \quad \text{– (Equation 6)}$$

where $a(i)$is the average distance between a sample and all other samples within the same cluster, and $b(i)$is the average distance between that sample and the nearest neighboring cluster. Silhouette values range from -1 to +1, with higher values indicating better cluster separation and stronger cluster cohesion.

The silhouette scores for both datasets are presented in **Figure 3(b)**. For the pure-oil dataset, the scores remained substantially higher across all tested cluster numbers than those obtained for the fried-food dataset. The pure-oil spectra produced silhouette values of 0.4631 ($k = 2$), 0.3309 ($k = 3$), 0.3331 ($k = 4$), 0.3434 ($k = 5$), 0.3474 ($k = 6$), 0.3294 ($k = 7$), 0.2007 ($k = 8$), and 0.2042 ($k = 9$). In contrast, the fried-food dataset generated considerably lower scores of 0.2163 ($k = 2$), 0.1526 ($k = 3$), 0.1271 ($k = 4$),

0.1255 ($k = 5$), 0.0626 ($k = 6$), 0.0574 ($k = 7$), 0.0625 ($k = 8$), and 0.0480 ($k = 9$). These values indicate stronger cluster compactness and greater separation among the pure-oil spectra, whereas the fried-food samples form weaker and more overlapping clusters.

Importantly, the quantitative clustering metrics closely mirror the patterns previously observed in the t-SNE visualizations. Sections 3.1 and 3.2 showed that the pure-oil dataset consistently formed compact and stable groupings across a wide range of perplexity values, while the fried-food dataset exhibited broader overlap and reduced class definition. The higher silhouette values obtained for the pure oils provide quantitative confirmation of these visual observations, demonstrating that the stronger separation seen in the t-SNE embeddings reflects genuine structure within the original Raman feature space rather than artifacts of dimensionality reduction. Conversely, the lower silhouette values for the fried-food dataset support the interpretation that matrix-induced spectral variability weakens the natural organization of the oil classes. In practical terms, higher WCSS reduction and higher silhouette scores indicate that the Raman spectra naturally organize into distinguishable groups, which generally creates a more favorable foundation for subsequent machine-learning classification.

Taken together, the elbow and silhouette analyses provide complementary evidence that the Raman spectra of pure oils possess a stronger intrinsic clustering tendency than those obtained from fried-food samples. The agreement between the t-SNE observations and the quantitative clustering metrics further supports the central premise of this study: the degree of spectral organization present within the data is closely linked to the effectiveness of subsequent machine-learning analysis.

### 3.4 K-Means Cluster Composition in Pure Oils and Fried-Food Samples

The K-means cluster assignments provide a direct way to evaluate how well the unsupervised model recovers the oil-class structure suggested by the t-SNE visualizations in **Figure 2** and **Figure S3** and by the clustering quality metrics in **Figure 3**. For the pure-oil dataset, the cluster composition shown in **Figure 4(a)** and summarized in **Table 1** indicates that the dominant class structure is recovered with high fidelity. **Table 1** reports the original numerical cluster identifiers (Clusters 0-4) generated by the K-means algorithm prior to the dominant-label mapping used for visualization in **Figure 4**.

It is important to note that **Figure 4** differs fundamentally from **Figure 2**. **Figure 2** displays the true oil labels assigned during sample collection, whereas **Figure 4** displays the K-means cluster assignments projected onto the same t-SNE coordinates. Thus, the degree of visual agreement between **Figure 2** and **Figure 4** provides a direct indication of how successfully the unsupervised algorithm recovers the natural class structure present in the Raman data.

The cross-tabulation in **Table 1** confirms this interpretation quantitatively. In the pure-oil dataset, GNO, SO, and VO are concentrated almost entirely within single dominant clusters, while PO and SOYO are distributed across a smaller number of clusters with some overlap. Although K-means does not produce a perfect one-to-one correspondence between clusters and oil classes, the observed overlap remains limited and is likely associated with similarities in lipid composition and unsaturation among chemically related oils. The strong visual correspondence between **Figure 2(a)** and **Figure 4(a)** further supports these findings. Regions occupied by GNO, SO, and VO in the true-label visualization remain largely preserved following K-means clustering, indicating that the dominant spectral organization identified by t-SNE is also recovered in the original high-dimensional feature space used for clustering.

A different pattern is observed for the fried-food dataset. As shown in **Figure 4(b)** and summarized in **Table 1**, the cluster structure is less compact, the class boundaries are broader, and the overlap among oil categories is more pronounced than in the pure-oil case. This agrees with the weaker visual separation seen in **Figure 2(b)** and the lower silhouette scores reported in **Figure 3(b)**. Unlike the pure-oil dataset, the agreement between **Figure 2(b)** and **Figure 4(b)** is noticeably weaker, indicating that the underlying oil-class structure is less clearly defined within the fried-food matrix.

The absence of a clearly isolated PO-dominant group in **Figure 4(b)** is a direct consequence of this weaker separability. Examination of the cluster-label assignments for the fried-food dataset shows that PO spectra are not concentrated in a single dominant cluster; instead, they are distributed across multiple clusters, with substantial mixing with other oil categories. Specifically, the cluster composition in **Table 1** shows that PO samples are divided primarily between Cluster 0 (80 spectra) and Cluster 4 (43 spectra), preventing PO from becoming the dominant class within any individual cluster. As a result, no separate PO cluster appears as a distinct dominant region in the **Figure 4(b)** visualization. This is not a plotting error, but rather a reflection of the fact that PO has lower cluster purity in the fried-food matrix and is therefore less cleanly resolved by K-means than in the pure-oil dataset.

The cluster-label comparison in **Table 1** further supports this interpretation. For the pure-oil dataset, the table shows strong concentration of class labels into dominant clusters, whereas in the fried-food dataset the same classes are spread more broadly across multiple clusters. This difference indicates that the intrinsic Raman fingerprint of each oil is more clearly preserved in the pure-oil system, whereas the food matrix introduces enough additional spectral variability to weaken the natural grouping of the samples. The overall trend is therefore fully consistent with the earlier t-SNE analysis in **Figure 2**, the perplexity stability analysis in **Figure S1** and **Figure S2**, and the silhouette-based comparison in **Figure 3**.

We performed data transformation even with MinMaxScaler to compare results with StandardScaler. MinMaxScaler did not improve the clustering performance [28]. One possible reason is that MinMaxScaler is sensitive to extreme values, which can compress the majority of observations into a narrow range. StandardScaler, by centering features around zero and scaling them according to their standard deviation, may better preserve the relative variation among features. Consequently, StandardScaler produced a clearer clustering structure, as indicated by the WCSS elbow pattern and higher Silhouette scores.

Taken together, the unsupervised clustering results show that the pure-oil spectra possess stronger intrinsic class organization than the fried-food spectra. The strong agreement among **Figure 2**, **Figure 3**, **Figure 4**, and **Table 1** establishes a coherent picture: the pure-oil samples exhibit compact, well-defined clusters, whereas the fried-food samples show broader overlap and less distinct class boundaries. This progression from qualitative visualization to quantitative clustering metrics and finally to cluster-label composition provides a consistent basis for the later supervised classification analysis.

### 3.5 Decision Tree Classification: Comparison Between Pure Oils and Fried-Food Samples

To systematically evaluate the influence of Decision Tree complexity on Raman-based edible-oil classification, three modeling strategies were applied sequentially to four datasets: pure oils, original chips, paper-subtracted chips, and paper-and-potato-subtracted chips. First, a baseline Decision Tree was established using the initial model configuration. Subsequently, pre-pruning was performed by optimizing selected tree-growth hyperparameters, including max_depth, max_leaf_nodes, and min_samples_split, to constrain model complexity during tree construction. Finally, post-pruning was performed by tuning the ccp_alpha, which controls the removal of decision branches after tree development. The optimized hyperparameter configurations for the baseline, pre-pruned, and post-pruned Decision Tree models across all four datasets are summarized in **Table S1** in the Supplementary Information.

The principal hyperparameters shown in **Table S1** have distinct roles in controlling Decision Tree complexity. ccp_alpha is the cost-complexity pruning parameter that penalizes tree complexity and removes branches that provide limited predictive benefit; max_depth specifies the maximum allowable depth of the tree; max_leaf_nodes limits the maximum number of terminal nodes; and min_samples_split defines the minimum number of samples required to split an internal node [28-29]. As summarized in **Table S1**, the pure-oil dataset achieved optimal pre-pruned performance using a relatively shallow tree (max_depth = 4), whereas the original chips and paper-subtracted datasets required greater tree depth (max_depth = 6) to accommodate the increased complexity introduced by the food matrix. The paper-and-potato-subtracted dataset again required a shallower tree (max_depth = 4), indicating that NNLS-based subtraction of the fitted food matrix-reference spectra simplified the underlying classification problem. The post-pruned models introduced non-zero ccp_alpha values for the three chips datasets (0.0115, 0.0153, and 0.0222, respectively), whereas the pure-oil dataset retained ccp_alpha = 0, consistent with the strong intrinsic separability of the pure-oil spectra. Overall, **Table S1** demonstrates that the optimal Decision Tree complexity depends strongly on the spectral characteristics and matrix complexity of each dataset, with physics-informed matrix subtraction generally enabling more compact and interpretable models.

### 3.5.1 Baseline Decision Tree Performance

To establish a reference point for subsequent model optimization, baseline Decision Tree classifiers were developed using the default scikit-learn parameters for the pure-oil dataset and for the three chips datasets representing progressively different levels of matrix contribution. The chips datasets consisted of the original Raman spectra, spectra after subtraction of the paper contribution using NNLS, and spectra after subtraction of both paper and potato contributions. Because identical train-test splits were maintained throughout the analysis, the resulting performance metrics provide a direct comparison of the influence of matrix effects and physics-informed preprocessing on model behavior.

All baseline Decision Tree models achieved perfect performance on their respective training datasets, yielding accuracy, precision, recall, and F1-scores of 100%. While these results demonstrate that the fully grown trees were sufficiently flexible to separate the training samples, such perfect training performance is also characteristic of Decision Tree overfitting, particularly when high-dimensional datasets are analyzed. Consequently, test-set performance provides a more meaningful measure of model generalization and practical predictive capability.

For the pure-oil dataset, the baseline Decision Tree maintained perfect performance on the independent test set, achieving 100% accuracy, precision, recall, and F1-score, as shown in **Figure 5(a)**. This outcome indicates that the Raman spectra of the five edible oils possess highly distinctive and well-separated chemical signatures. The result is consistent with the strong class organization observed earlier in the t-SNE visualizations and K-means clustering analyses, where the oil classes formed compact and well-defined groups with minimal overlap. A markedly different behavior was observed for the chips' datasets. The original chips spectra produced a test accuracy of only 62.59%, despite perfect training performance (**Figure 5(b)**). This substantial train-test gap demonstrates that the classifier was able to memorize the training data but struggled to generalize to previously unseen spectra. Such behavior is consistent with the increased spectral complexity introduced by the food matrix, which weakens class separability and produces more ambiguous decision boundaries.

Application of NNLS-based matrix subtraction significantly improved classifier generalization. Removal of the paper contribution increased test accuracy from 62.59% to 77.04%, representing an improvement of approximately 14 %, as shown in **Figure 5(c)**. This considerable increase suggests that paper-derived Raman contributions introduce substantial spectral variability that masks oil-specific information relevant for classification. When both paper and potato contributions were subtracted, the baseline classifier achieved a test accuracy of 73.70% (**Figure 5(d)**). Although slightly lower than the paper-subtracted dataset, performance remained substantially higher than that obtained using the original chips spectra.

These results demonstrate that matrix-related spectral contributions play a major role in limiting classification performance. More importantly, the observed improvements were achieved without changing the classifier architecture, training strategy, or hyperparameters. The only difference among the chips analyses was the spectral representation generated through NNLS-based preprocessing. Consequently, the improved performance can be attributed to enhanced spectral separability arising from physics-informed matrix removal rather than from increased model complexity. Collectively, these findings support the central premise of this study that the intrinsic structure of the Raman spectral space strongly influences machine-learning success and that Physics-Informed AI approaches can improve the extraction of chemically meaningful information from complex food matrices. A consolidated comparison of the accuracy, precision, recall, and F1-scores obtained from the baseline, pre-pruned, and post-pruned Decision Tree models across all datasets is provided in **Table 2**.

### 3.5.2 Pre-Pruned Decision Tree Analysis

To reduce overfitting while maintaining interpretability, pre-pruning was applied by constraining tree growth through optimization of the maximum tree depth, maximum number of leaf nodes, and minimum number of samples required for node splitting. The resulting test-set confusion matrices for the pure-oil dataset, original chips dataset, paper-subtracted chips dataset, and paper- plus potato-subtracted chips dataset are presented in **Figure 6(a-d)**, respectively. The corresponding models were selected using five-fold cross-validation on the training data to improve generalization performance while preventing excessive model complexity.

The pure-oil dataset continued to exhibit excellent classification performance following pre-pruning, achieving a test accuracy of 100%. The corresponding confusion matrix (**Figure 6a**) shows complete separation among all five oil classes, indicating that the Raman spectra of the pure oils possess sufficiently strong class-specific signatures to permit perfect classification even after substantial reduction in model complexity. This result demonstrates that the discriminatory information present in the pure-oil spectra is highly robust and does not depend on a highly complex decision-tree structure.

In contrast, the original chips dataset showed only a modest improvement relative to the baseline Decision Tree. Test accuracy increased from 62.6% to 64.8%, indicating that pre-pruning reduced some degree of overfitting but could not fully overcome the limitations imposed by matrix-induced spectral overlap (**Figure 6b**). The confusion matrix continued to exhibit substantial misclassification among several oil classes, suggesting that the food matrix weakens the separability of the oil-specific Raman signatures and creates more ambiguous classification boundaries.

More pronounced improvements were observed after application of NNLS-based matrix subtraction. For the paper-subtracted dataset, the pre-pruned classifier achieved a test accuracy of 73.7% (**Figure 6c**), while the paper- and potato-subtracted dataset achieved 76.7% (**Figure 6d**). Relative to the original chips spectra, these results demonstrate that removal of matrix contributions improves class discrimination and enables the pruned trees to construct more effective decision boundaries using fewer spectral variables. The reduced confusion among several oil classes further suggests that NNLS preprocessing enhances the visibility of oil-related Raman features that are otherwise partially masked by food-matrix signals.

Comparison with the baseline Decision Tree results reveals several important trends. For pure oils, pre-pruning maintained perfect classification performance while simultaneously producing a substantially simpler and more interpretable model, indicating that the spectral classes are intrinsically well separated. For the original chips' dataset, the improvement was limited, highlighting the difficulty of classifying spectra strongly influenced by matrix interference. However, the matrix-corrected datasets demonstrated substantially better performance than the untreated chips spectra, confirming that improved spectral representation contributes more to generalization performance than increasing model complexity alone. Collectively, these findings indicate that the effectiveness of Decision Tree classification is governed not only by classifier design but also by the underlying spectral structure of the dataset. The NNLS-based preprocessing therefore serves as an important physics-informed step that improves the quality of the feature space available to the classifier.

The optimized pre-pruned Decision Tree developed for the pure-oil dataset provides a highly interpretable classification model while maintaining perfect test-set performance. The graphical structure of the tree and the corresponding feature-importance analysis are shown in **Figure 7(a-b)**. Remarkably, the classifier achieved complete discrimination of all five edible oils using only four Raman variables out of the original 1866 spectral features. Thus, only approximately 0.21% of the available spectral variables were required for accurate classification, demonstrating that most of the discriminatory information is concentrated within a very small subset of the Raman spectrum. This pronounced reduction is consistent with feature-selection strategies reported elsewhere in the Frugal AI literature for high-dimensional spectral and hyperspectral data, where combinatorial and Markov-decision-process-based approaches are used to identify a small subset of maximally informative variables from a much larger measurement space [26].

To further test the hypothesis that the discriminatory information required for pure-oil classification is concentrated within these four Raman variables, a new dataset was constructed using only the four selected features. Remarkably, this reduced dataset independently achieved 100% test-set accuracy, confirming that the complete 1866-feature spectral representation is not necessary for accurate classification. This result provides direct evidence of the Frugal AI potential of the proposed approach, in which predictive performance is retained while substantially reducing data dimensionality and resource requirements. When represented as Pandas DataFrames, the four-feature dataset occupied approximately ~ 81 KB, compared with ~ 14.3 MB for the original 1866-feature dataset, corresponding to a memory requirement of only approximately 0.56% of the full dataset. Thus, the four-feature representation reduced the data footprint by approximately 99.44% while maintaining perfect classification accuracy of 100% [results are in pre-pruned decision tree section of the notebook in 28]. This combination of high predictive performance and minimal data requirements highlights the potential of the approach for resource-constrained, Edge AI, and embedded Raman-sensing applications.

To better understand **Figure 7(a)**, it is useful to briefly examine how Decision Trees perform classification. Decision Trees operate by recursively partitioning the dataset using threshold values applied to selected Raman variables. At each node, the algorithm identifies the Raman band and threshold that provide the greatest reduction in class uncertainty. The parameter ‘samples’ indicate the number of spectra reaching a given node, while ‘value’ represents the distribution of spectra among the five oil classes. The ‘Gini index’ quantifies the degree of class mixing within a node, with larger values indicating greater heterogeneity and a value of zero indicating that all spectra belong to a single class. Consequently, the objective of the Decision Tree algorithm is to progressively reduce Gini impurity through a sequence of spectral decisions until highly homogeneous or completely pure terminal nodes are obtained. In practical terms, the tree can be viewed as a series of simple "if-then" Raman-based decisions that progressively reduce uncertainty regarding oil identity and ultimately produce the final classification.

**Figure 7(a)** illustrates the decision pathway of the optimized pre-pruned tree, in which classification of the five oils is achieved through a short sequence of intensity-based thresholds applied to the four selected Raman variables. At the root node, spectra are first partitioned according to the normalized intensity at ~ 1649 $cm^{-1}$: samples with intensity greater than 0.94 are classified directly as SO, reflecting an exceptionally strong and distinctive signal at this band, while samples at or below this threshold proceed to further splitting. Within this branch, the tree next evaluates intensity at ~ 1322 $cm^{-1}$: values at or below 0.01 lead to a subsequent split at ~ 1127 $cm^{-1}$, where intensities below -0.42 are classified as PO and intensities above this value are classified as VO. When intensity at ~ 1322 $cm^{-1}$ instead exceeds 0.01, a final threshold at ~

1273 $cm^{-1}$ separates GNO (intensity $\leq 0.71$) from SOYO (intensity $> 0.71$). Each terminal node reaches complete class purity (Gini = 0), with all training samples correctly isolated by class, underscoring the sharpness of these intensity thresholds as diagnostic markers for pure-oil identification.

Feature-importance analysis (**Figure 7(b)**) revealed that the classification process was governed almost equally by four Raman bands located near ~ 1127, ~ 1273, ~ 1322, and ~ 1649 $cm^{-1}$. These represent complementary physicochemical descriptors of edible oils [4-5]. The ~1649 $cm^{-1}$ band reflects the degree of unsaturation through C=C stretching vibrations (more double bonds produce stronger signal), whereas the ~1273 $cm^{-1}$ band is associated with cis double-bond environments (arrangement of double bonds in a chain) and therefore captures differences in lipid fluidity and molecular packing. The ~1322 $cm^{-1}$ band reflects $CH_2$ twisting and chain conformational order (Frying alters the chain conformation), while the ~1127 $cm^{-1}$ band represents C-C skeletal stretching and hydrocarbon-chain organization. The vibrational mode at 1127 $cm^{-1}$ involves the vibration of carbon backbone itself and hence it is present even in fried samples too. Collectively, these variables describe unsaturation, double-bond geometry, chain conformation and carbon-skeleton order, providing a chemically meaningful basis for discrimination among the five oils.

Feature-importance analysis (**Figure 7(b)**) reveals that the four selected Raman variables contribute almost equally to the classification process, indicating that the Decision Tree relies on a balanced combination of complementary spectral descriptors rather than a single dominant feature. The band near 1649 $cm^{-1}$, assigned to C=C stretching vibrations, forms the root node of the tree and plays a major role in the early separation of sunflower oil, consistent with its stronger unsaturation-related spectral signature [4]. The Raman variable near 1322 $cm^{-1}$ subsequently separates groundnut and soybean oils from palm and vanaspati oils. Based on the spectral assignments reported by Silverstein and Webster, this band may be associated with O-H related vibrations and could reflect differences in naturally occurring minor constituents such as phenolic compounds and antioxidants among the oils [31]; however, it may also capture broader variations in lipid-chain conformation and molecular organization such as $CH_2$ twisting and hydrocarbon-chain conformation [4-5]. Within this branch, the 1273 $cm^{-1}$ feature further discriminates groundnut oil from soybean oil. This band may be associated with C-O-C vibrations of alkyl aryl ether groups and therefore could be influenced by differences in naturally occurring tocopherol (vitamin E) related constituents [31], while also reflecting variations in the local molecular environment of unsaturated lipid species including cis =C-H deformation and cis-double-bond environment [4]. Finally, the Raman variable near 1127 $cm^{-1}$ differentiates palm oil from vanaspati oil. Although this spectral region has been associated with secondary alcohol related vibrations [31], it lies within the fingerprint region of C-C skeletal stretching [4-5, 32-34] and is therefore best interpreted as a discriminative marker capturing subtle differences in overall molecular composition and hydrocarbon-chain organization between these oils. Together, these four nearly equally weighted features explain the balanced importance distribution observed in **Figure 7(b)** and provide a chemically interpretable basis for the perfect classification achieved by the Decision Tree model.

An important observation is that the four variables contribute nearly equally to the final model, with each accounting for approximately 25% of the total importance. This balanced importance distribution suggests that oil discrimination is not driven by a single dominant Raman marker but rather by complementary chemical information distributed across several characteristic lipid vibrations. The ability to achieve perfect classification using only four highly informative Raman bands highlights the strong intrinsic spectral separability of the pure-oil dataset and demonstrates the potential for developing ultra-compact Raman-based classification systems with minimal computational requirements. This balanced importance distribution suggests that the four Raman variables provide a near mutually exclusive yet collectively exhaustive (MECE) description of the discriminatory lipid chemistry. Rather than relying on a dominant spectral marker, the classifier integrates complementary information related to unsaturation, antioxidant content, and skeletal organization (evident from their finger print region), together capturing the principal compositional dimensions separating the five oils.

A markedly different behavior was observed for the fried-food datasets. Unlike the pure-oil classifier, which relied on only four Raman variables, the chips classifiers required substantially larger numbers of spectral features to achieve satisfactory performance. This increased complexity reflects the presence of matrix-derived spectral contributions originating from the potato substrate and paper, which partially obscure oil-specific Raman signatures. Consequently, the Decision Trees must utilize a broader set of spectral variables to construct effective classification boundaries.

For the original chips' dataset, the pre-pruned model contained approximately 29 non-zero important features, whereas the number decreased to 15 after subtraction of the paper contribution and to 11 after subtraction of both paper and potato contributions. The progressive reduction in feature count accompanied by improved classification performance provides strong evidence that the NNLS-based preprocessing enhances spectral interpretability by removing non-informative matrix signals and concentrating discriminatory information into a smaller subset of Raman variables. The corresponding feature-importance plots are presented in **Figure 8(a-c)**, while the complete Decision Tree structures are provided in the Supplementary Information **(Figures S4-S6)** and accompanying Jupyter notebooks for readers interested in the detailed classification pathways and threshold values used by the models.

### 3.5.3 Post-Pruned Decision Trees Analysis

To further improve model generalization while maintaining interpretability, cost-complexity post-pruning was applied to the optimized Decision Tree models. Unlike pre-pruning, which restricts tree growth during model construction, post-pruning begins with a fully developed tree (from optimized pre-pruned decision tree from **section 3.5.2**) and subsequently removes branches that contribute little to predictive performance. This procedure reduces model complexity while preserving the most informative decision boundaries. The resulting confusion matrices for the pure-oil dataset, original chips dataset, paper-subtracted chips dataset, and paper- plus potato-subtracted chips dataset are shown in **Figure 9(a-d)**, respectively.

To determine the optimal level of post-pruning, cost-complexity pruning was applied using the complexity parameter (*ccp_alpha*). Cost-complexity pruning introduces a penalty for tree size and therefore balances classification performance against model complexity. First, the cost-complexity pruning path was generated using the training data, producing a sequence of candidate $\alpha$ values and their corresponding leaf-node impurities. As $\alpha$ increases, the Decision Tree is progressively simplified by removing branches that contribute relatively little to classification performance. The final $\alpha$ value in the pruning path was excluded because it produces an extremely pruned, trivial tree containing only a single root node and no meaningful decision branches. The remaining candidate *ccp_alpha* values were evaluated using GridSearchCV with five-fold stratified cross-validation. For each candidate $\alpha$ value, a new Decision Tree was constructed and its performance was assessed using weighted recall. The pruning trajectory was further monitored through changes in total leaf impurity, tree depth, and number of nodes. As $\alpha$ increased, progressively weaker branches were removed, generally reducing tree depth and node count while increasing overall leaf impurity. The optimal $\alpha$ value was selected as the one producing the highest mean cross-validated weighted recall, thereby selecting the pruning level that provided the best classification performance without retaining unnecessary tree complexity and increasing the risk of overfitting.

Weighted recall was selected as the optimization metric because it provides a direct measure of how effectively the classifier recovers spectra belonging to each oil class. In multiclass edible-oil classification, incorrectly assigning a sample to another oil category results in a false-negative error for the true class and a false-positive error for the predicted class. A classifier that achieves high weighted recall therefore minimizes missed identifications across all oil categories while accounting for class frequencies within the dataset. Because the primary objective of this study was reliable recognition of all oil types rather than optimization of a single class, weighted recall provided a robust criterion for selecting the final post-pruned model. The recall-versus-α analysis further enabled direct comparison of training and test performance, allowing identification of the pruning level that produced the strongest generalization to previously unseen Raman spectra

For the pure-oil dataset, post-pruning maintained perfect classification performance. The confusion matrix shown in **Figure 9(a)** exhibits complete separation among all five oil classes, resulting in accuracy, precision, recall, and F1-score values of 100%. The identical performance observed for the baseline, pre-pruned, and post-pruned models demonstrates that the Raman spectra of pure oils possess exceptionally strong intrinsic class structure. Consequently, highly accurate classification can be achieved irrespective of tree complexity, provided that the key discriminative Raman variables are retained.

The original chips dataset exhibited only a modest response to post-pruning. Classification accuracy increased slightly relative to the baseline model (to 64.4%) but remained comparable to the pre-pruned classifier (64.8%) (**Figure 9(b)**). Significant class overlap and misclassification persisted, indicating that pruning alone cannot fully compensate for the reduced spectral separability introduced by the food matrix. These results suggest that the primary limitation is not excessive tree complexity but rather the presence of matrix-derived Raman signals that obscure the oil-specific signatures.

A substantially different pattern emerged for the NNLS-processed datasets. For the paper-subtracted chips spectra, post-pruning slightly increased classification accuracy to 79.25 %, representing a significant improvement over both the baseline model (77.0%) and the pre-pruned model (73.7%) (**Figure 9(c)**). Similarly, for the paper- and potato-subtracted dataset, post-pruning achieved an accuracy of 79.63 %, exceeding both the baseline (73.7%) and pre-pruned (76.7%) classifiers (**Figure 9(d)**). The corresponding confusion matrices show noticeably improved class discrimination and reduced misclassification relative to earlier models.

Comparison of **Figures 5, 6,** and **9** reveals several important trends. First, classifier optimization has minimal influence on the pure-oil dataset because the spectral classes are already perfectly separated. Second, for the untreated chips spectra, improvements achieved through pruning alone remain limited because matrix interference continues to dominate the classification problem. Third, and most importantly, the combination of NNLS-based matrix subtraction and post-pruning produces the strongest overall performance. Once physically meaningful matrix contributions are removed, the classifier can construct substantially more effective decision boundaries while simultaneously avoiding overfitting.

These findings demonstrate that improvements in predictive performance arise not simply from modifying the machine-learning algorithm but from enhancing the underlying spectral representation itself. The strong gains observed for the NNLS-corrected datasets therefore support the central premise of this study: machine-learning success is governed primarily by the quality and separability of the Raman spectral information available to the classifier. Physics-informed preprocessing improves that spectral structure, while post-pruning enables the classifier to exploit it more effectively, resulting in improved generalization and higher predictive accuracy. The overall progression in classification performance from the baseline to the optimized pre-pruned and post-pruned models is summarized in **Table 2**, which highlights the substantial gains obtained after NNLS-based matrix correction.

The post-pruned Decision Tree models were further examined to identify the Raman variables responsible for classification and to determine how post-pruning influenced model complexity relative to the pre-pruned classifiers. By eliminating branches that contributed minimally to predictive performance, post-pruning produced more compact and interpretable models while maintaining or improving classification accuracy. The resulting feature-importance analyses are presented in **Figure 10** for the pure-oil dataset and **Figure 11** for the chips' datasets. The post-pruned decision tree pathways for chips, paper-subtracted chips, and paper-and potato-subtracted chips are shown in **Figure S7-S9**.

For the pure-oil dataset, the post-pruned classifier retained the same four dominant Raman variables identified in the pre-pruned model, namely bands near ~ 1127, ~ 1273, ~ 1322, and ~ 1649 $cm^{-1}$ (**Figure 10a-b**). These four variables were sufficient to achieve perfect classification of all five edible oils, despite the original Raman spectra containing 1866 spectral features. Thus, only approximately 0.21% of the available spectral variables were required for accurate prediction. The persistence of these same four Raman bands across both pre-pruned and post-pruned models demonstrates their robustness as discriminatory spectral markers and confirms that the intrinsic spectral structure of the pure-oil dataset is highly stable.

More substantial changes were observed for the chips' datasets. The post-pruned classifier developed using the original chips spectra relied on approximately 16 important Raman variables, representing a considerable reduction relative to the 29 variables required by the corresponding pre-pruned model. Despite this reduction in complexity, classification performance remained comparable, indicating that many of the variables selected by the larger pre-pruned tree contributed little additional predictive information. This observation suggests that post-pruning successfully eliminated weak decision branches associated with noise and matrix-related variability.

The benefits of post-pruning became even more pronounced after NNLS-based matrix subtraction. For the paper-subtracted chips dataset, the number of important variables decreased from 15 in the pre-pruned model to only 5 in the post-pruned model, while test accuracy increased substantially from 73.7% to 79.3%. Similarly, for the paper- and potato-subtracted dataset, the number of important variables decreased from 11 to only 4, while test accuracy increased from 76.7% to 79.6%. These results demonstrate that NNLS preprocessing not only improves classification accuracy but also concentrates discriminatory information into a remarkably small number of Raman variables.

A noteworthy observation is that two Raman variables near ~1652 and ~1126 $cm^{-1}$ were consistently retained in both NNLS-corrected datasets. These bands correspond closely to the ~1649 and ~1127 $cm^{-1}$ markers identified in the pure-oil models and are respectively associated with lipid unsaturation, antioxidant content and fingerprint region [4-5]. The persistence of these features across pure-oil and matrix-containing datasets indicates that they represent robust oil-specific signatures that remain largely unaffected by food-matrix contributions. Moreover, their combined importance increased from approximately 50% in the pure-oil model to about 62% in the paper-subtracted model and approximately 89% in the paper-plus-potato-subtracted model, suggesting that progressive matrix removal concentrates classification around the most fundamental descriptors of edible-oil chemistry.

Additional variables appearing in the NNLS-corrected models provide insight into the influence of frying and residual matrix effects. The ~1258 and ~1629 $cm^{-1}$ bands are located within unsaturation-sensitive spectral regions and may reflect thermally modified double-bond environments, while the ~860 $cm^{-1}$ feature disappears after potato subtraction, suggesting a contribution from residual oil-matrix interactions [4-5]. After removal of both paper and potato contributions, the classifier introduces a Raman variable near ~ 2919 $cm^{-1}$ corresponding to CH stretching vibrations of lipid hydrocarbon chains, indicating that high-wavenumber lipid information becomes accessible once matrix interference is minimized.

A particularly noteworthy finding is that the matrix-corrected chips datasets ultimately required only four to five dominant Raman variables, approaching the level of simplicity observed for the pure-oil classifier. This trend suggests that matrix-derived spectral contributions are responsible for much of the apparent complexity observed in the original chips' dataset. Once these contributions are removed through NNLS decomposition, the underlying oil-specific Raman signatures become more prominent, enabling highly compact and interpretable classification models.

Overall, comparison of the pre-pruned and post-pruned models reveals a consistent trend toward reduced model complexity following NNLS-based preprocessing. While the original chips dataset required many spectral variables to compensate for matrix interference, the paper-subtracted and paper-plus-potato-subtracted datasets achieved higher predictive performance using only a handful of Raman features. These findings further support the central hypothesis of this work: improving the physical representation of the Raman spectra through Physics-Informed AI can be more beneficial than increasing classifier complexity. The resulting models are not only more accurate but also more interpretable and computationally efficient, making them attractive candidates for future Frugal AI, Edge AI, and embedded Raman-sensing applications.

### 3.5.4 Implications of Pre- and Post-Pruning

The pre-pruning and post-pruning analyses collectively provide important insight into the relationship between spectral structure, model complexity, and classification performance. For the pure-oil dataset, both pruning approaches achieved perfect classification while consistently identifying the same four Raman variables near ~ 1127, ~ 1273, ~ 1322, and ~ 1649 $cm^{-1}$ as the dominant discriminatory features. The remarkable stability of these variables across different optimization strategies indicates that the pure-oil spectra possess strong intrinsic class structure and that the information required for classification is concentrated within a very small portion of the Raman spectrum. In practical terms, perfect discrimination of five edible oils was achieved using only four variables from an original feature space of 1866 Raman shifts. The performance metrics summarized in **Table 2** further demonstrate that improvements in classification accuracy were accompanied by reductions in model complexity, particularly for the NNLS-corrected chips datasets.

From a Frugal AI perspective, this result is particularly significant because accurate classification was achieved using less than 0.21% of the available spectral variables. The ability to dramatically reduce feature dimensionality while maintaining predictive performance demonstrates that effective Raman-based classification does not necessarily require large feature sets or highly complex models. Instead, a compact set of chemically meaningful spectral variables can provide sufficient information for reliable decision-making. This behavior mirrors feature-selection approaches described for other high-dimensional spectral and hyperspectral modalities, where the central objective is to identify a minimal set of maximally discriminative variables from a much larger feature space [26].

A contrasting behavior was observed for the food-matrix datasets. The original chips spectra required substantially more spectral variables and exhibited only modest improvements following pruning, reflecting the difficulty of separating oil-specific information from matrix-derived spectral contributions. Although pruning reduced model complexity, classification performance remained limited by the reduced separability of the underlying spectral data. These results suggest that optimization of the learning algorithm alone cannot fully compensate for the effects of matrix interference.

The benefits of Physics-Informed AI became evident after NNLS-based matrix subtraction. For both the paper-subtracted and paper-plus-potato-subtracted datasets, post-pruning not only improved classification accuracy but also dramatically reduced the number of important Raman variables compared with the corresponding pre-pruned models. In particular, the paper-subtracted model required only five important features, while the paper-plus-potato-subtracted model relied on just four features, approaching the simplicity of the pure-oil classifier. The simultaneous increase in predictive performance and reduction in model complexity indicates that NNLS successfully concentrates discriminatory spectral information by removing non-informative matrix contributions.

Overall, the pruning experiments demonstrate that classification performance is governed primarily by the quality and organization of the spectral information rather than by model complexity alone. The combination of NNLS-based physics-informed preprocessing and Decision Tree optimization produced models that were more accurate, more interpretable, and substantially more compact. These findings support the broader concepts of Explainable AI (XAI), Frugal AI, and Edge AI, where reliable decision-making is achieved using a minimal set of physically meaningful variables and reduced computational resources.

An important observation is that the analytical workflow adopted in this study follows a MECE (Mutually Exclusive, Collectively Exhaustive) structure. The t-SNE and K-means analyses quantify intrinsic spectral organization, NNLS provides physics-informed decomposition of matrix contributions, Decision Trees evaluate predictive discrimination, and Frugal-AI-inspired feature reduction identifies the minimum set of informative Raman variables. Because each stage addresses a distinct analytical objective without redundancy, while collectively covering the complete progression from raw spectra to deployment-oriented classification, the framework provides a coherent and interpretable characterization of Raman-based edible-oil classification.

## 4. Conclusions and Future Outlook

This study developed an integrated Raman spectroscopy and machine-learning framework combining intrinsic spectral structure, classification performance, and Physics-Informed Artificial Intelligence (PI-AI) for edible-oil analysis. Building on our earlier Raman-chemometric marker framework [1] and Raman-machine-learning classification study [2], the present work combined t-SNE, elbow and silhouette analyses, K-means clustering, interpretable Decision Trees, and PI-AI to investigate classification behavior in pure oils and complex fried-food matrices. Unsupervised analyses showed stronger spectral organization in pure oils, with greater cluster compactness, class purity, and separability than fried-food samples, where matrix effects caused substantial spectral overlap. For pure oils, optimized Decision Trees achieved perfect classification using only four Raman variables near ~ 1127, ~ 1273, ~ 1322, and ~ 1649 $cm^{-1}$. Despite 1866 original Raman variables, only approximately 0.21% of the feature space was required to discriminate the five edible oils, with the same four variables independently selected by both pre-pruned and post-pruned models. Furthermore, two of these spectral markers, located near ~1650 and ~1127 $cm^{-1}$, persisted in the NNLS-corrected food-matrix datasets and became increasingly dominant following matrix subtraction, indicating that lipid unsaturation and hydrocarbon-chain skeletal organization (C–C stretching) are the most robust Raman descriptors for edible-oil discrimination under both ideal and complex food-matrix conditions.

Beyond classification performance, the present framework exhibits a MECE (Mutually Exclusive, Collectively Exhaustive) architecture. Unsupervised learning, physics-informed decomposition, supervised classification, and frugal feature selection each contribute unique and non-overlapping analytical information while collectively providing a complete description of the Raman-based classification problem. This MECE organization strengthens interpretability, reduces analytical redundancy, and offers a systematic blueprint for future Physics-Informed AI workflows in spectroscopy and food-quality monitoring.

A major contribution was the integration of PI-AI through Non-Negative Least Squares (NNLS)-based spectral decomposition, using the physical constraint that Raman contributions are non-negative and additive. NNLS estimated reference-aligned matrix contributions from sampling paper and potato components, substantially improving machine-learning performance. Optimized post-pruned classifiers achieved accuracies of nearly 80% for the paper-subtracted and paper-plus-potato-subtracted datasets, compared with approximately 64.4% for the original chips' spectra. Model complexity also decreased substantially, with important Raman variables reduced from 29 in the pre-pruned original-chips model to five and four variables in the corresponding NNLS-corrected datasets. These findings demonstrate the value of combining PI-AI with interpretable Decision Trees and support Frugal AI principles, where accurate predictions use minimal data and computational resources [26, 27], making the approach suitable for portable Raman, Edge AI, and embedded platforms. This behavior mirrors feature-selection approaches described for other high-dimensional spectral and hyperspectral modalities, where the central objective is to identify a minimal set of maximally discriminative variables from a much larger feature space [26].

Overall, the results demonstrate that classification performance depends strongly on the quality and organization of spectral information, with NNLS-based preprocessing often providing greater improvement than classifier optimization alone. The framework therefore aligns with XAI, PI-AI, and Frugal AI principles by prioritizing physically meaningful, interpretable, and computationally efficient representations. Future research should extend the approach to additional food matrices, adulterated and recycled frying oils, blended formulations, and samples with different frying conditions, storage histories, and processing environments, while exploring ensemble learning, feature-attribution methods, lightweight deep learning, and real-time embedded analytics. Collectively, Raman spectroscopy, unsupervised spectral analysis, NNLS-based PI-AI, and interpretable machine learning provide a pathway toward robust, transparent, computationally efficient, and field-deployable food-quality monitoring systems.

**Acknowledgements:** The authors are grateful to the founder Chancellor and acknowledge the Management of Sri Sathya Sai Institute of Higher Learning, Andhra Pradesh, India, for facilitating this research. The authors are also thankful to the Central Research Instrumentation Facility (SSSIHL-CRIF), Prasanthi Nilayam, for providing the necessary instrumental support.

**Author contributions:** A.S.: Experiments and Raman spectra collection, sample preparation, conceptualization, methodology, investigation, software, data curation, formal analysis, visualization, and approval of final version of manuscript. C.S.N.: software, data curation, formal analysis, visualization, review and editing, the concept of NNLS, and approval of final version of manuscript. S.M.V.: supervision, review and editing, and approval of final version of manuscript. J.G.: conceptualization, supervision, review and editing, and approval of final version of manuscript. D.L.N.K.: Ideation, software, formal analysis, writing, reviewing and editing, the code for decision tree and Physics-Informed AI, and approval of final version of manuscript.

**Funding:** The authors did not receive any financial support from any funding agencies for the submitted work.

**Data Availability:** The datasets and analysis scripts generated and used in the present study are available from the corresponding author(s) upon reasonable request and will be shared on a case-by-case basis, subject to ongoing related research activities and institutional policies. These are hosted on GitHub [28].

**Declarations:** Competing Interests The authors declare no competing interests.

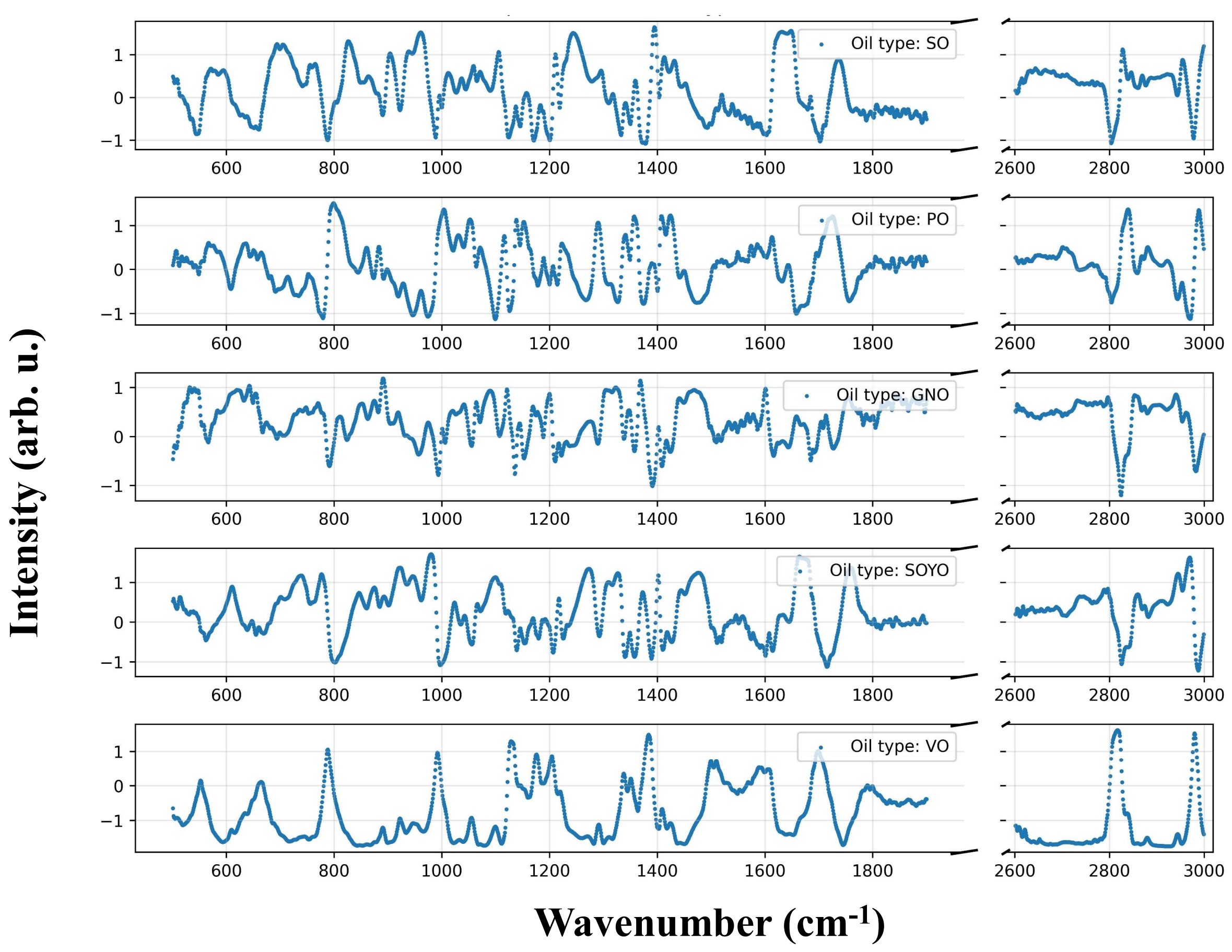

**Figure 1:** Mean Raman spectra of the five edible oils investigated in this study (SO, PO, GNO, SOYO, and VO). A broken x-axis is used to omit the spectral region between 1900 and 2600 cm⁻¹, where no relevant Raman bands were observed. Differences in spectral intensity and peak distribution indicate compositional variations among the oils and provide the basis for subsequent machine-learning classification. X-axis represents features/wavenumbers.

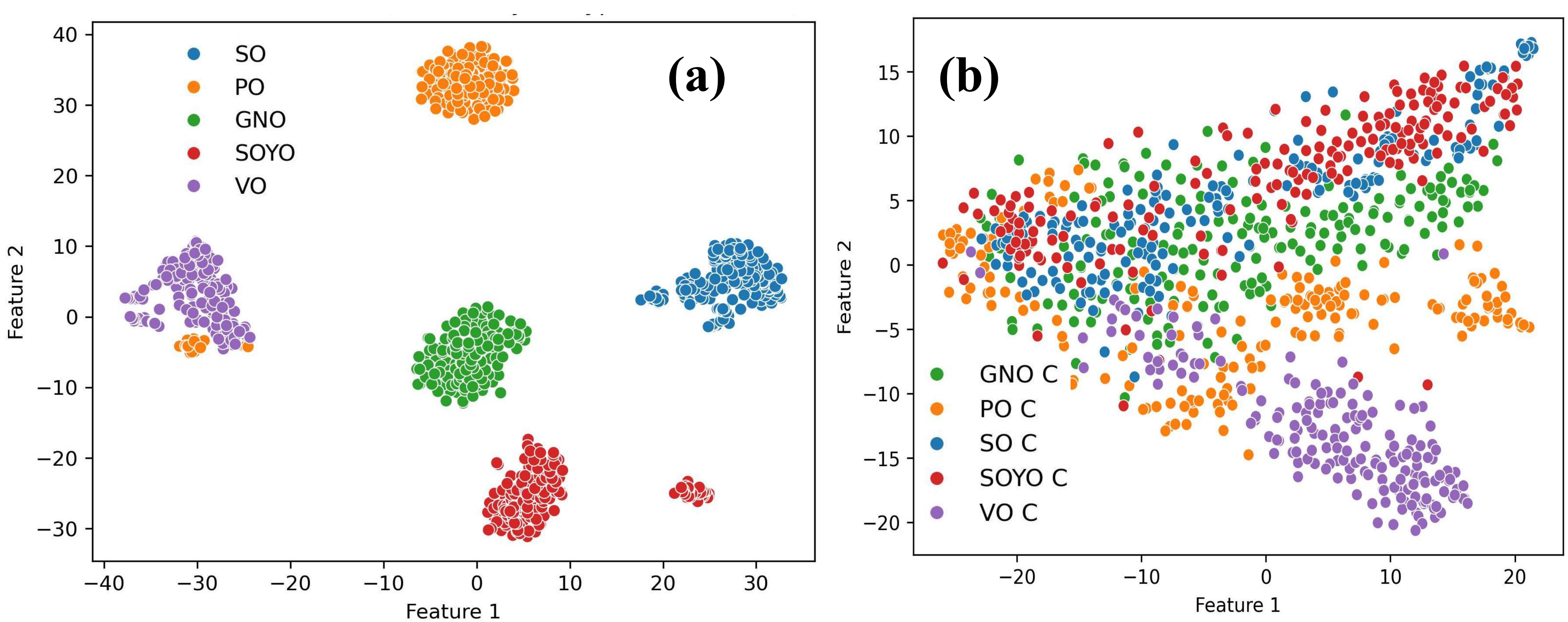

**Figure 2:** Two-dimensional t-SNE embeddings of Raman spectra for **(a)** pure edible oils and **(b)** fried-food samples. In the pure-oil dataset, SO, VO, and GNO form relatively compact and distinct regions, while PO and SOYO show partial proximity; in the fried-food dataset, the same classes become broader and more overlapping, indicating reduced separability due to matrix effects. The default perplexity is 30.

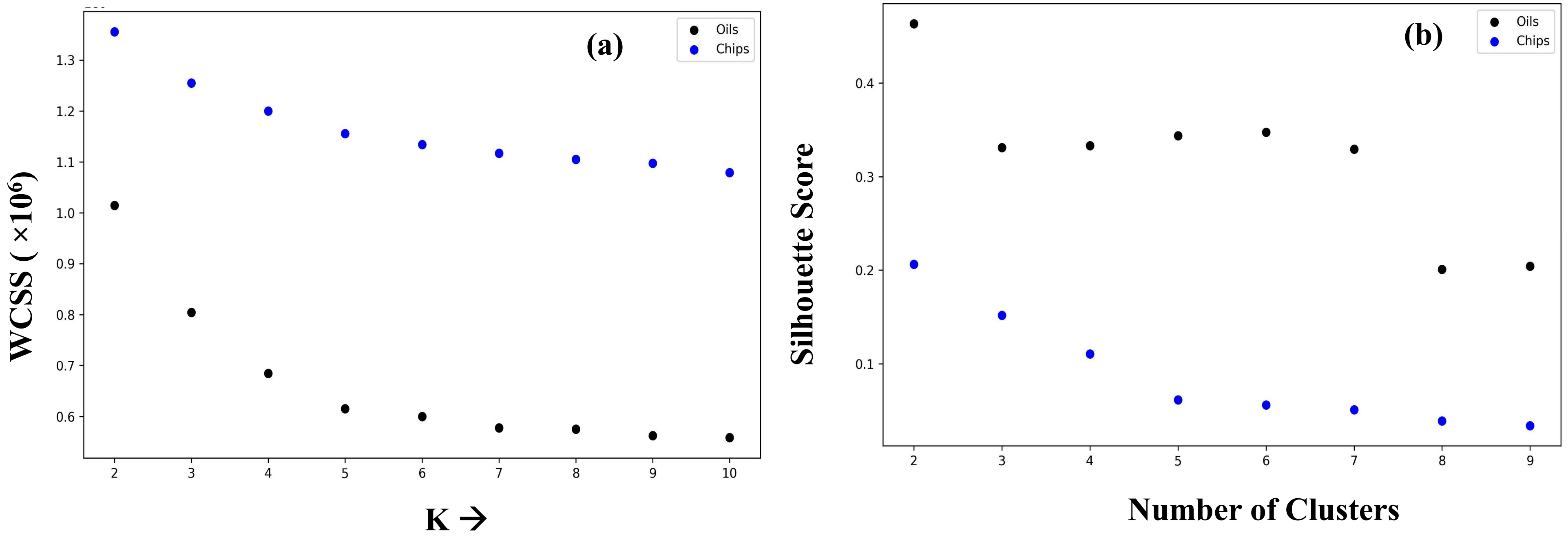

**Figure 3.** Unsupervised clustering quality assessment for the pure-oil and fried-food (chips) Raman spectral datasets. **(a)** Elbow plots showing the variation of within-cluster sum of squares (WCSS) as a function of the number of K-means clusters for the pure-oil (black) and chips (blue) datasets. In both cases, WCSS decreases with increasing cluster number; however, the pure-oil dataset exhibits a more pronounced elbow near the expected five-cluster solution, indicating stronger underlying class organization. **(b)** Silhouette scores obtained for K-means clustering across different cluster numbers for the same datasets. The consistently higher silhouette scores observed for the pure-oil spectra (black) demonstrate greater cluster compactness and inter-cluster separation, whereas the lower scores for the chips' dataset (blue) indicate increased class overlap and reduced separability arising from food-matrix contributions. Together, these results quantitatively confirm that the intrinsic spectral structure of pure oils is more distinct than that of matrix-containing fried-food samples.

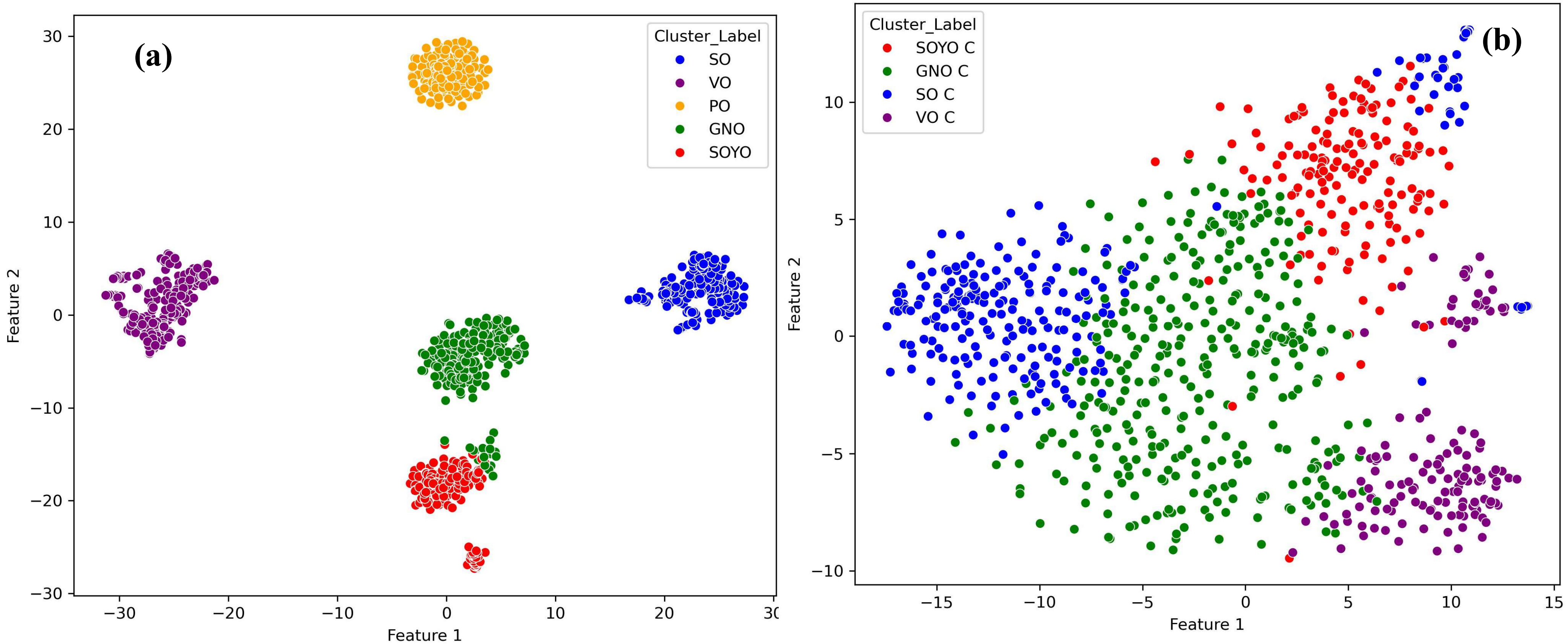

**Figure 4:** K-means cluster assignments projected onto the t-SNE embeddings shown in **Figure 2**. **(a)** For pure oils, the K-means clusters closely reproduce the true-label structure, indicating strong intrinsic class organization. **(b)** For fried-food samples, greater overlap is observed and agreement with the true-label organization is reduced. PO does not appear as a dominant cluster because its spectra are distributed primarily between Clusters 0 and 2 rather than concentrated within a single majority cluster.

| | Pure Oils | | | | | | Food Matrix | | | | | |
|---|---|---|---|---|---|---|---|---|---|---|---|---|
| | **0** | **1** | **2** | **3** | **4** | **All** | **0** | **1** | **2** | **3** | **4** | **All** |
| **GNO** | 200 | 0 | 0 | 0 | 0 | 200 | 96 | 0 | 50 | 0 | 34 | 180 |
| **PO** | 0 | 40 | 160 | 0 | 0 | 200 | 90 | 7 | 7 | 33 | 43 | 180 |
| **SO** | 0 | 0 | 0 | 200 | 0 | 200 | 49 | 21 | 38 | 0 | 72 | 180 |
| **SOYO** | 18 | 0 | 0 | 0 | 182 | 200 | 46 | 17 | 66 | 0 | 51 | 180 |
| **VO** | 0 | 200 | 0 | 0 | 0 | 200 | 62 | 0 | 0 | 112 | 6 | 180 |
| **Total** | 218 | 240 | 160 | 200 | 182 | 1000 | 343 | 45 | 161 | 145 | 206 | 900 |

**Table 1:** Cross-tabulation between true oil labels and K-means cluster assignments for the pure-oil and fried-food datasets. Clusters 0-4 represent the original numerical cluster identifiers produced by the K-means algorithm prior to cluster relabelling for visualization in **Figure 5**. The table summarizes cluster purity and class overlap and explains why PO does not emerge as a dominant cluster in **Figure 5(b)**.

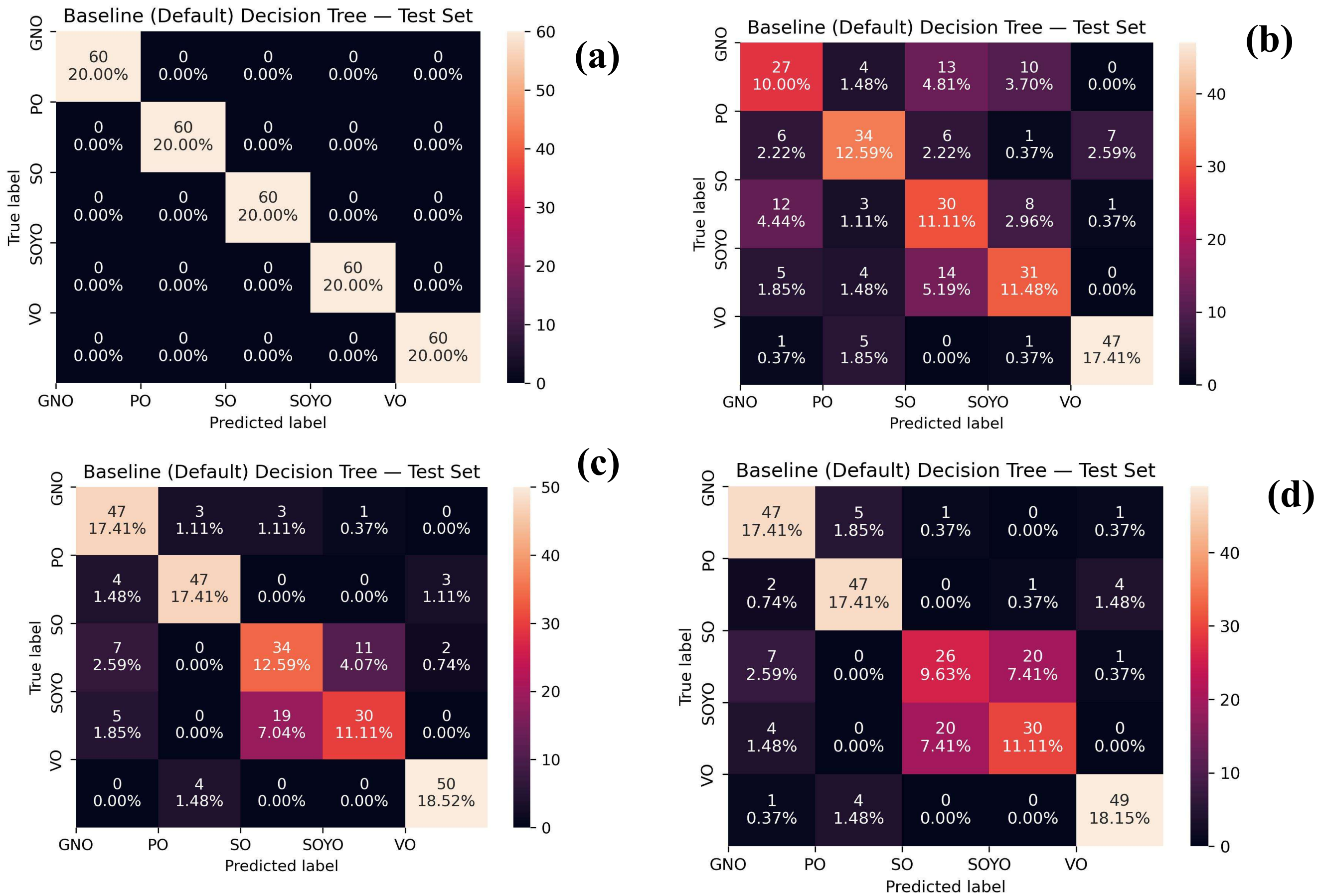

**Figure 5.** Test-set confusion matrices obtained using the baseline (default) Decision Tree classifier for **(a)** pure oils, **(b)** original chips spectra, **(c)** paper-subtracted chips spectra, and **(d)** paper- and potato-subtracted chips spectra. The corresponding accuracy, recall, precision, and F1-scores were 1.000, 1.000, 1.000, and 1.000 for pure oils; 0.6259, 0.6259, 0.6296, and 0.6268 for the original chips' dataset; 0.7704, 0.7704, 0.7694, and 0.7669 after paper subtraction; and 0.7370, 0.7370, 0.7284, and 0.7315 after paper and potato subtraction, respectively. The perfect classification achieved for pure oils contrasts sharply with the reduced performance observed for food-matrix samples, and NNLS-based matrix subtraction improved test performance relative to the original chips' spectra, demonstrating the benefit of physics-informed preprocessing for recovering oil-specific Raman information.

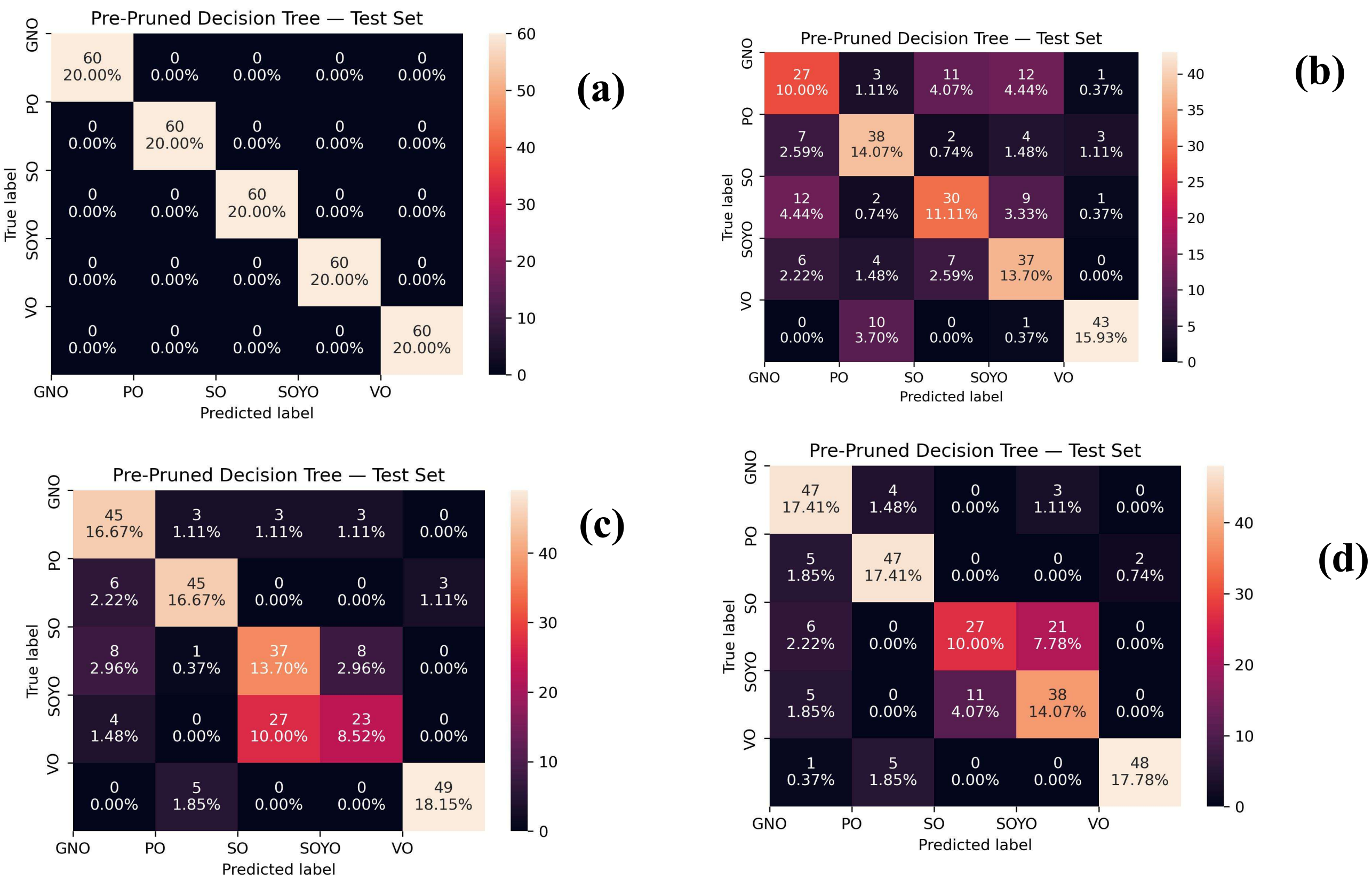

**Figure 6.** Test-set confusion matrices obtained using the optimized pre-pruned Decision Tree classifier for **(a)** pure oils, **(b)** original chips spectra, **(c)** paper-subtracted chips spectra, and **(d)** paper- and potato-subtracted chips spectra. The corresponding accuracy, recall, precision, and F1-scores were 1.000, 1.000, 1.000, and 1.000 for pure oils; 0.6481, 0.6481, 0.6538, and 0.6493 for the original chips' dataset; 0.7370, 0.7370, 0.7437, and 0.7323 for the paper-subtracted dataset; and 0.7667, 0.7667, 0.7714, and 0.7633 for the paper- and potato-subtracted dataset, respectively. Compared with the baseline Decision Tree results shown in **Figure 5**, pre-pruning maintained perfect classification of pure oils while improving generalization performance for the original chips and paper- plus potato-subtracted datasets. The results demonstrate that controlling tree complexity can reduce overfitting and enhance predictive performance, particularly when combined with physics-informed NNLS-based matrix correction.

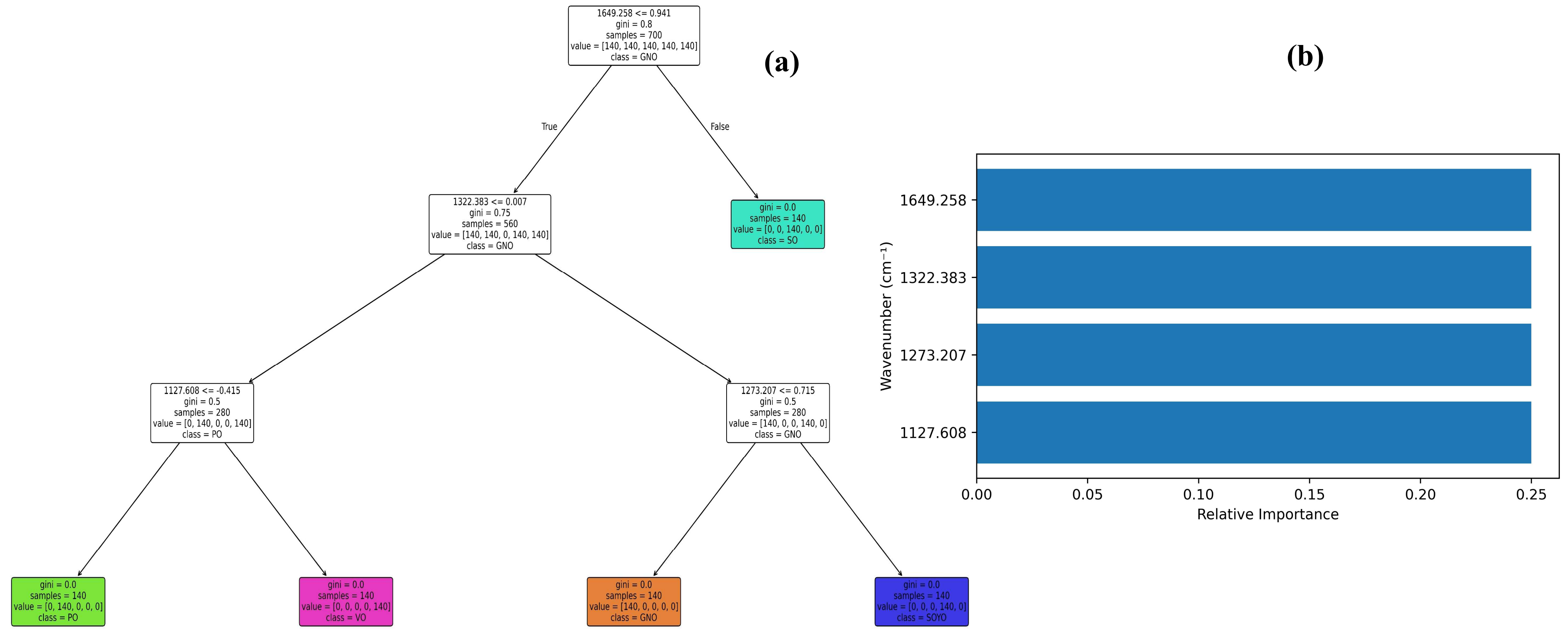

**Figure 7.** Interpretable pre-pruned Decision Tree model developed for the pure-oil Raman dataset.**(a)** Optimized pre-pruned Decision Tree showing the decision pathways used to classify the five edible oils. **(b)** Corresponding feature-importance analysis identifying the four Raman variables that govern classification. Despite the availability of 1866 spectral features, the model achieved perfect test-set classification using only four Raman bands (1127, 1273, 1322, and 1649 $cm^{-1}$), corresponding to approximately **0.21%** of the original feature space. The nearly equal importance of these variables indicates that oil discrimination is achieved through complementary spectral information associated with lipid-chain structure and unsaturation-related Raman vibrations.

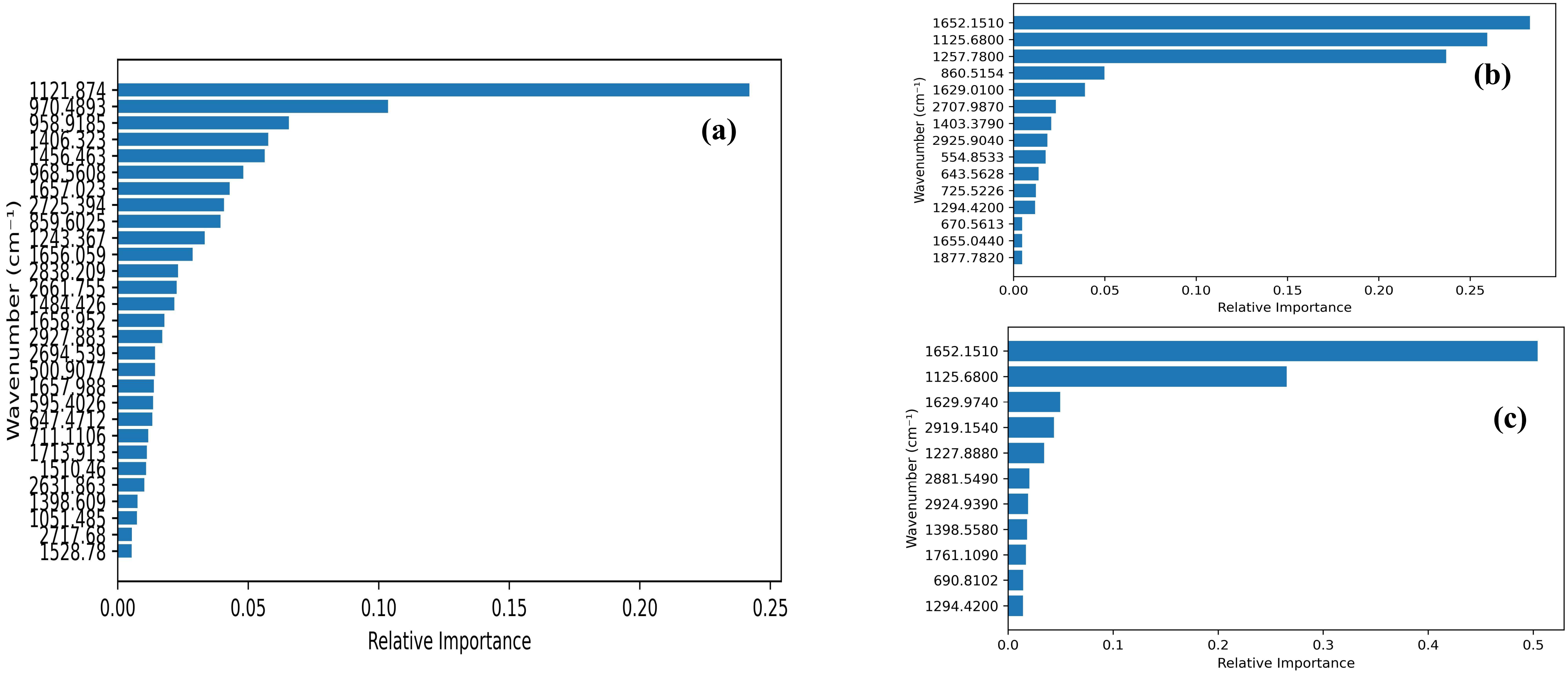

**Figure 8.** Feature-importance profiles obtained from the optimized pre-pruned Decision Tree models for **(a)** original chips spectra, **(b)** paper-subtracted chips spectra, and **(c)** paper- and potato-subtracted chips spectra. The original chips model required approximately 29 important Raman variables to classify the oils, whereas NNLS-based matrix correction reduced this number to 15 after paper subtraction and 11 after simultaneous paper and potato subtraction. The progressive reduction in the number of important variables, accompanied by improved classification performance, demonstrates that removal of matrix-derived spectral contributions concentrates discriminatory information into a smaller set of oil-specific Raman features. Complete Decision Tree structures and decision rules are provided in the Supplementary Information (**Figures S4-S6**) and accompanying Jupyter notebooks.

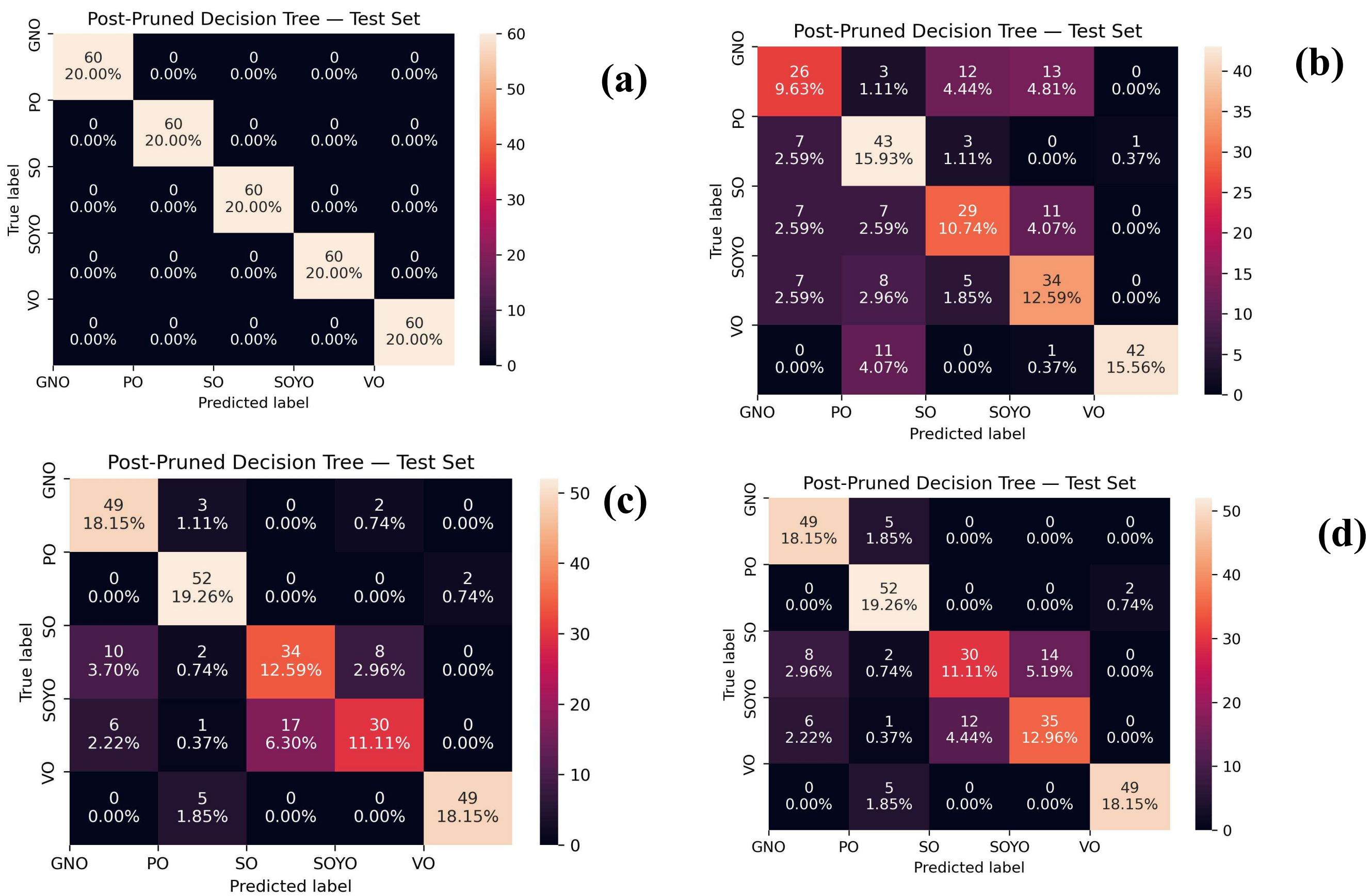

**Figure 9.** Test-set confusion matrices obtained using the optimized post-pruned Decision Tree classifier for **(a)** pure oils, **(b)** original chips spectra, **(c)** paper-subtracted chips spectra, and **(d)** paper- and potato-subtracted chips spectra. The corresponding accuracy, recall, precision, and F1-scores were 1.000, 1.000, 1.000, and 1.000 for pure oils; 0.6444, 0.6444, 0.6449, 0.6438 for the original chips' dataset; 0.8630, 0.8630, 0.8639, and 0.8619 for the paper-subtracted dataset; and 0.8540, 0.8540, 0.8551, and 0.8528 for the paper- and potato-subtracted dataset, respectively. Compared with the baseline (**Figure 5**) and pre-pruned (**Figure 6**) models, post-pruning produced the highest classification performance for the NNLS-processed datasets while maintaining perfect classification for pure oils. The results demonstrate that combining physics-informed matrix correction with post-pruning improves model generalization by reducing overfitting and preserving the most informative decision boundaries for oil discrimination.

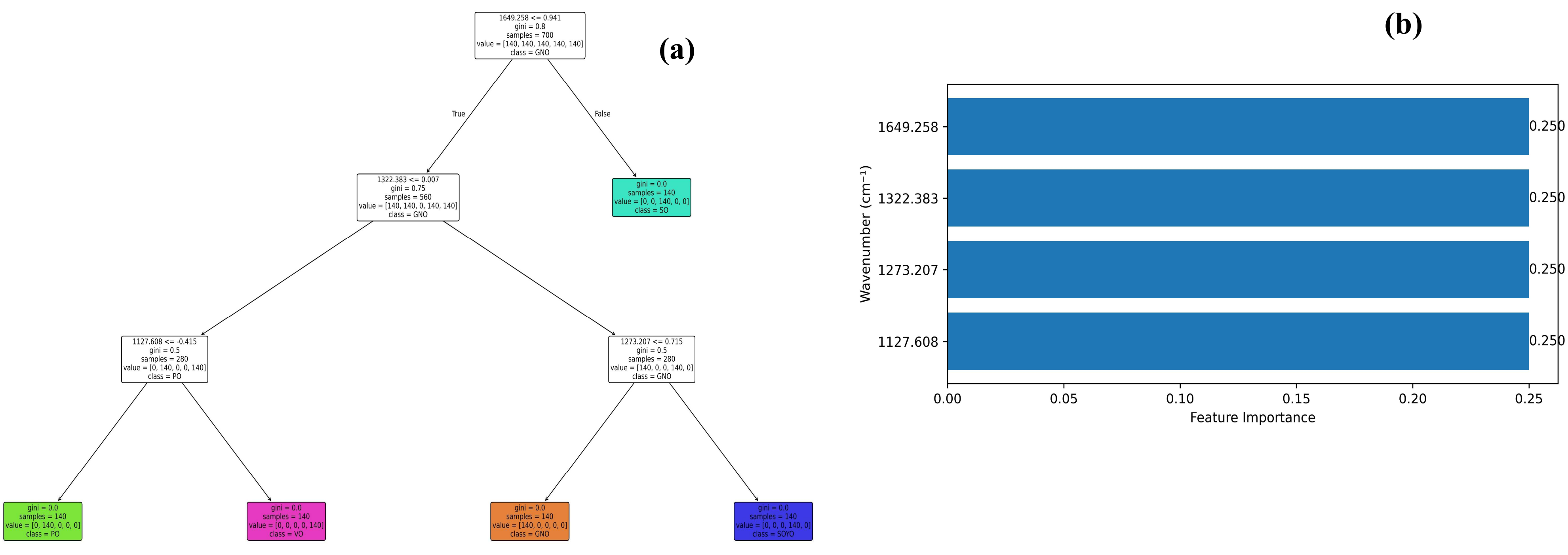

**Figure 10.** Interpretable post-pruned Decision Tree model developed for the pure-oil Raman dataset. **(a)** Optimized post-pruned Decision Tree showing the final classification pathway for the five edible oils. **(b)** Corresponding feature-importance analysis. The same four Raman variables (1127, 1273, 1322, and 1649 cm$^{-1}$) identified by the pre-pruned model were retained after post-pruning, demonstrating the robustness of these spectral markers. Perfect classification was achieved using only four variables from the original 1866-feature dataset, corresponding to approximately 0.21% of the available spectral information.

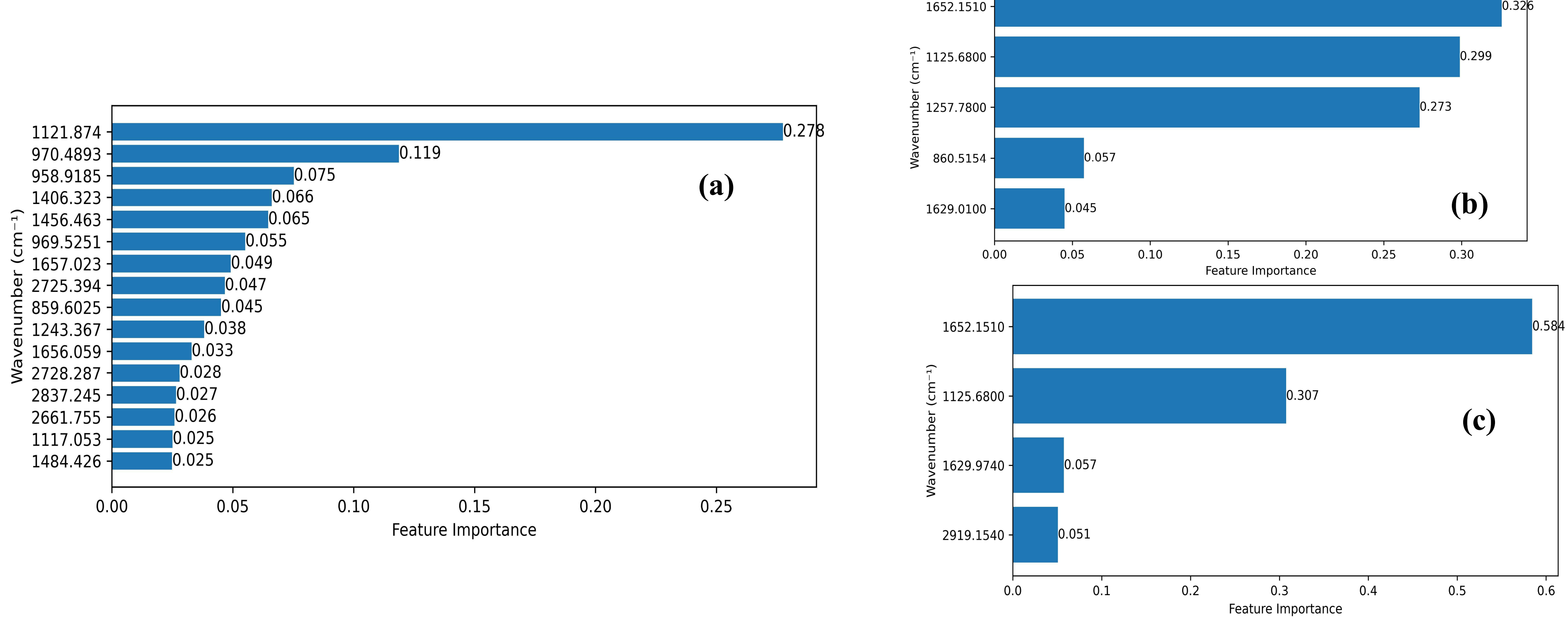

**Figure 11.** Feature-importance profiles obtained from the optimized post-pruned Decision Tree models for **(a)** original chips spectra, **(b)** paper-subtracted chips spectra, and **(c)** paper- and potato-subtracted chips spectra. Compared with the corresponding pre-pruned models, post-pruning reduced the number of important Raman variables from approximately 29 to 16 for the original chips' dataset, from 15 to 5 for the paper-subtracted dataset, and from 11 to 4 for the paper- and potato-subtracted dataset. The simultaneous reduction in model complexity and improvement in classification accuracy demonstrates that NNLS-based matrix correction concentrates discriminatory information into a smaller set of oil-specific Raman features, enabling more compact and interpretable classifiers.

| S. No. | Metric | Default | Pre-pruning | Post-pruning |
|---|---|---|---|---|
| **Pure Oils** | | | | |
| 1 | Accuracy | 1 | 1 | 1 |
| 2 | Recall | 1 | 1 | 1 |
| 3 | Precision | 1 | 1 | 1 |
| 4 | F1 Score | 1 | 1 | 1 |
| **Chips** | | | | |
| 1 | Accuracy | 0.6259 | 0.6481 | 0.6444 |
| 2 | Recall | 0.6259 | 0.6481 | 0.6444 |
| 3 | Precision | 0.6295 | 0.6538 | 0.6591 |
| 4 | F1 Score | 0.6267 | 0.6493 | 0.6456 |
| **Chips (Subtracted Paper Contribution)** | | | | |
| 1 | Accuracy | 0.7703 | 0.7370 | 0.8635 |
| 2 | Recall | 0.7703 | 0.7370 | 0.8635 |
| 3 | Precision | 0.7694 | 0.7437 | 0.8609 |
| 4 | F1 Score | 0.7669 | 0.7323 | 0.8617 |
| **Chips (Subtracted Paper and Potato Contribution)** | | | | |
| 1 | Accuracy | 0.737 | 0.7666 | 0.8539 |
| 2 | Recall | 0.737 | 0.7666 | 0.8539 |
| 3 | Precision | 0.7284 | 0.7714 | 0.8521 |
| 4 | F1 Score | 0.7314 | 0.7633 | 0.8516 |

**Table 2:** Summary of metrics for different datasets (Pure Oils, Chips, NNLS-corrected chips that include paper-subtracted chips, and paper-and-potato-subtracted chips)

| Dataset | Decision Tree | ccp_alpha | max_depth | max_leaf_nodes | min_samples_split |
|---|---|---|---|---|---|
| Oils | Default | 0 | None | None | 2 |
| | Pre-pruned | 0 | 4 | 50 | 10 |
| | Post-pruned | 0 | None | None | 2 |
| Chips | Default | 0 | None | None | 2 |
| | Pre-pruned | 0 | 6 | 50 | 10 |
| | Post-pruned | 0.0115 | None | None | 2 |
| Paper-subtracted | Default | 0 | None | None | 2 |
| | Pre-pruned | 0 | 6 | 50 | 70 |
| | Post-pruned | 0.0153 | None | None | 2 |
| Paper-and potato-subtracted | Default | 0 | None | None | 2 |
| | Pre-pruned | 0 | 4 | 50 | 10 |
| | Post-pruned | 0.0222 | None | None | 2 |

**Table S1:** Summary of Decision Tree hyperparameter configurations for the four datasets—pure oils, original chips, paper-subtracted chips, and paper-and-potato-subtracted chips—across the baseline, pre-pruned, and post-pruned models. The baseline model serves as the initial reference configuration, pre-pruning optimizes max_depth, max_leaf_nodes, and min_samples_split to control tree growth, and post-pruning tunes the cost-complexity parameter ccp_alpha to reduce unnecessary model complexity. The remaining hyperparameters were default and constant across all datasets and model configurations: class_weight = balanced, criterion = gini, max_features = None, min_impurity_decrease = 0, min_samples_leaf = 1, min_weight_fraction_leaf = 0, monotonic_cst = None, random_state = 0, and splitter = best.

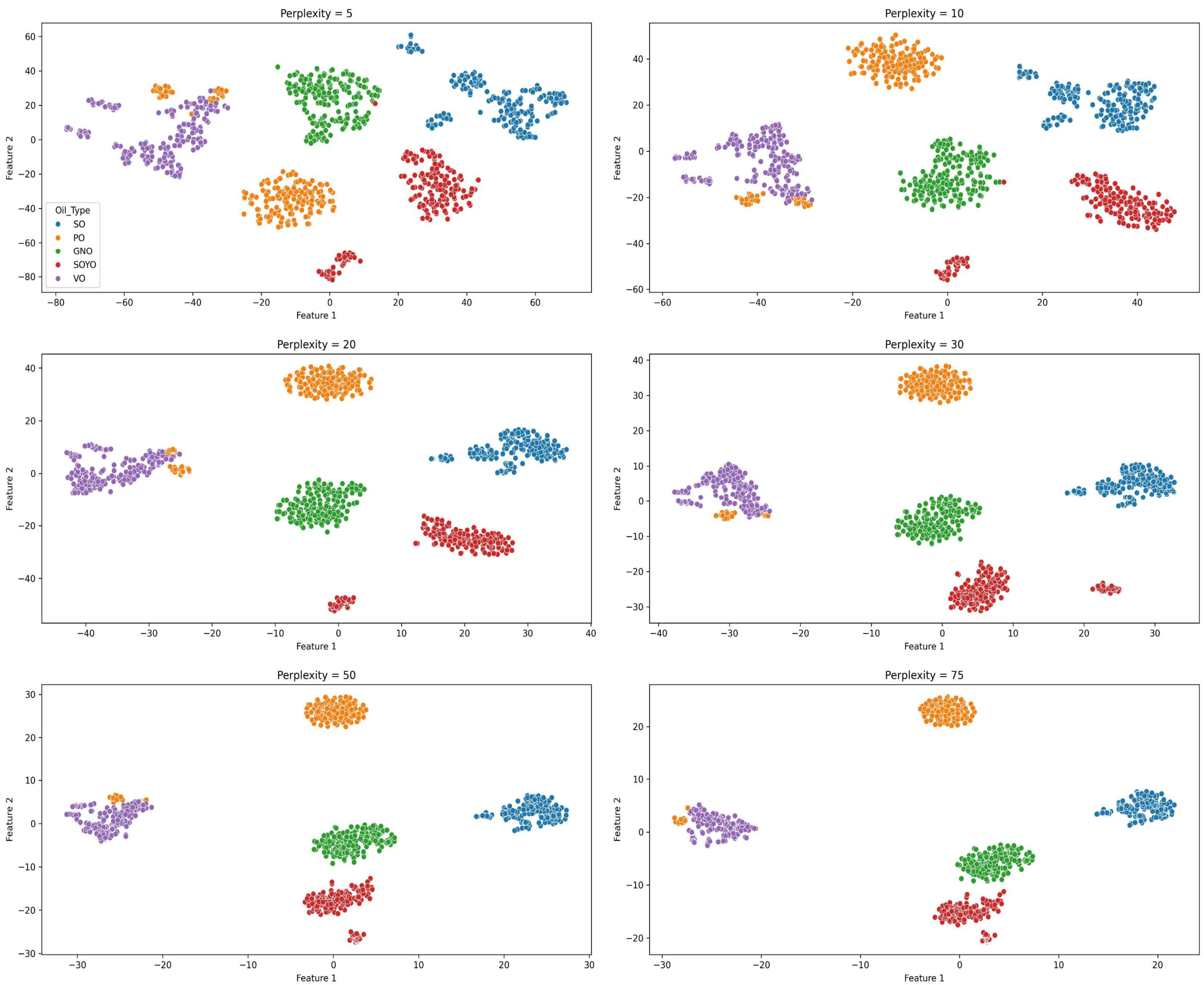

**Figure S1:** Effect of perplexity on the t-SNE visualization of pure-oil Raman spectra. Embeddings generated using perplexities of 5, 10, 20, 30, 50, and 75 show that the overall class organization remains stable across a broad range of neighborhood sizes, indicating robust underlying spectral structure.

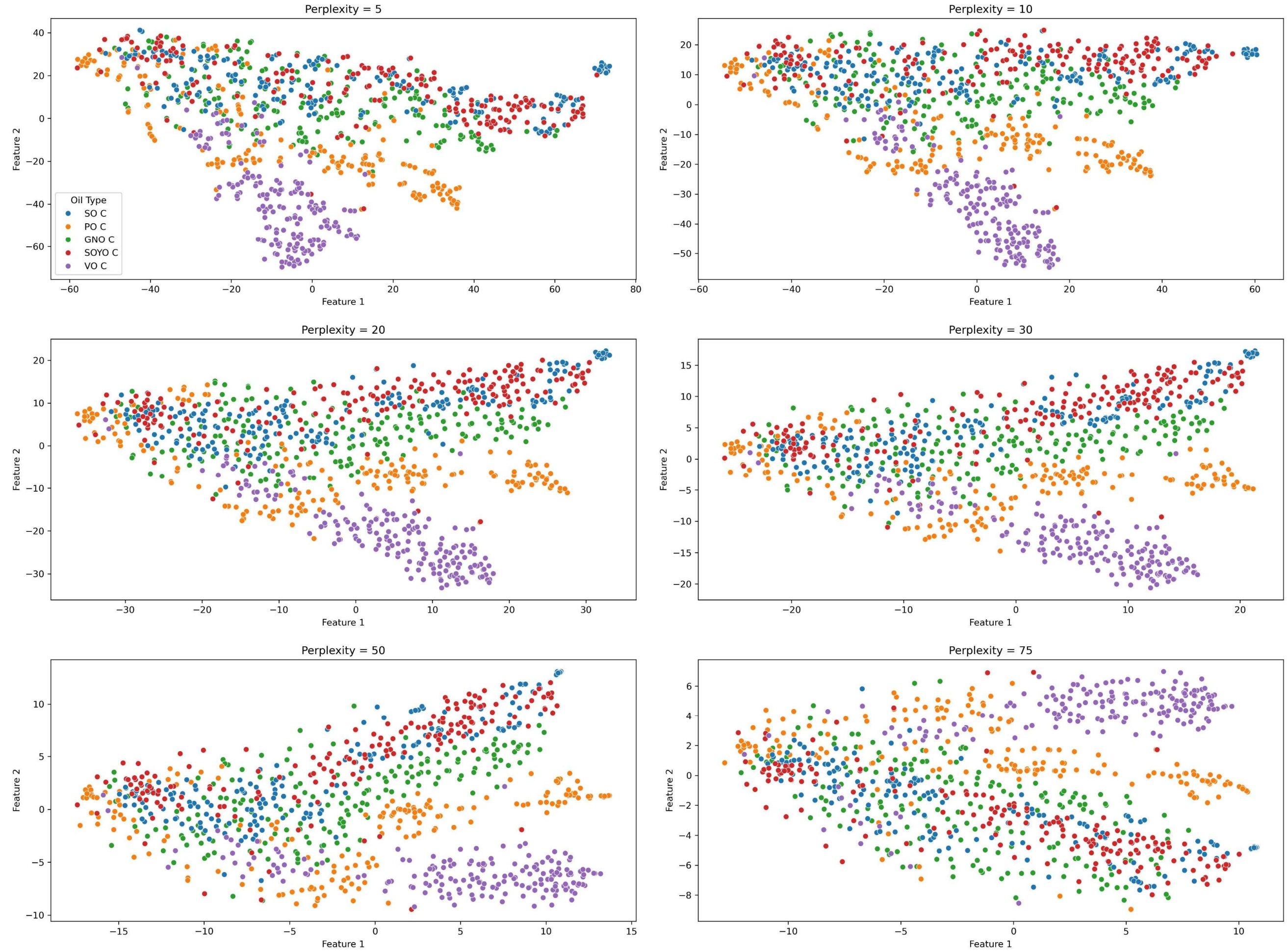

**Figure S2:** Effect of perplexity on the t-SNE visualization of fried-food Raman spectra. Although class-associated regions remain observable across all perplexity values, the embeddings consistently exhibit greater overlap and reduced compactness than the pure-oil dataset, reflecting increased spectral complexity introduced by the food matrix.

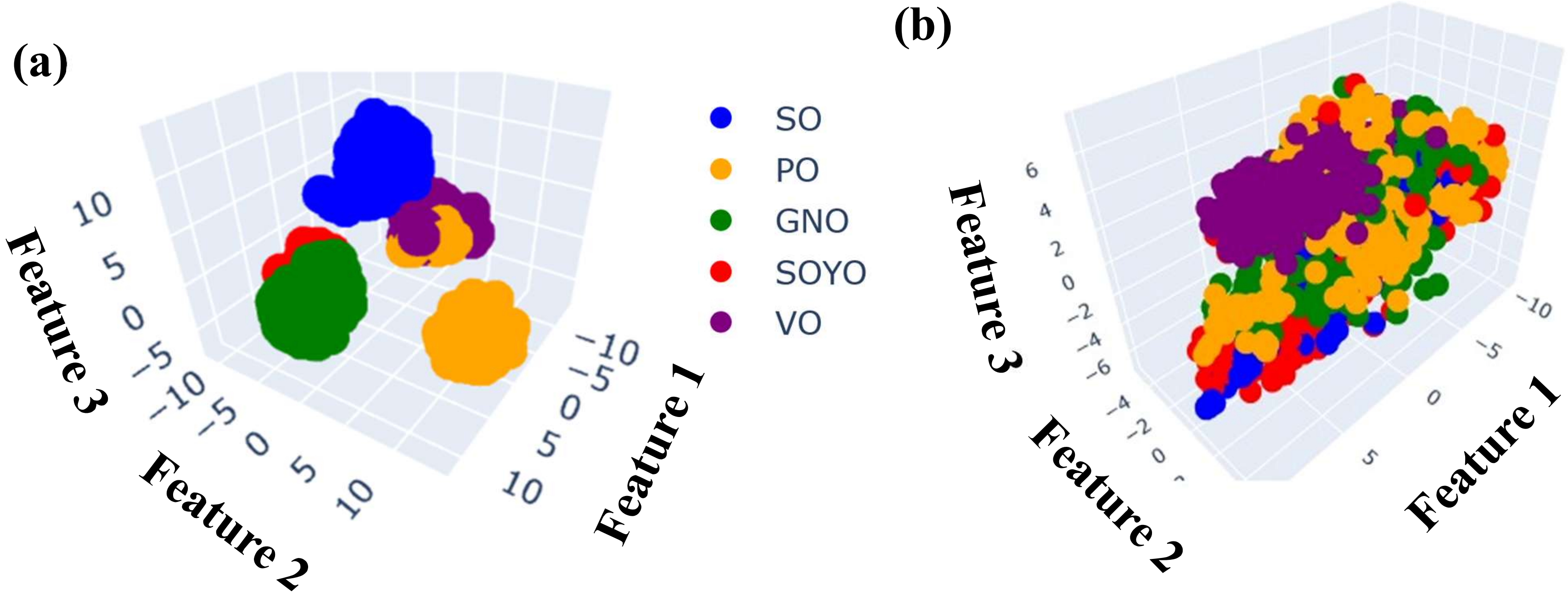

**Figure S3:** Three-dimensional t-SNE embeddings of Raman spectra for **(a)** pure oils and **(b)** fried-food samples. The 3D representation confirms the stronger class organization and separability observed in the pure-oil dataset relative to the fried-food matrix.

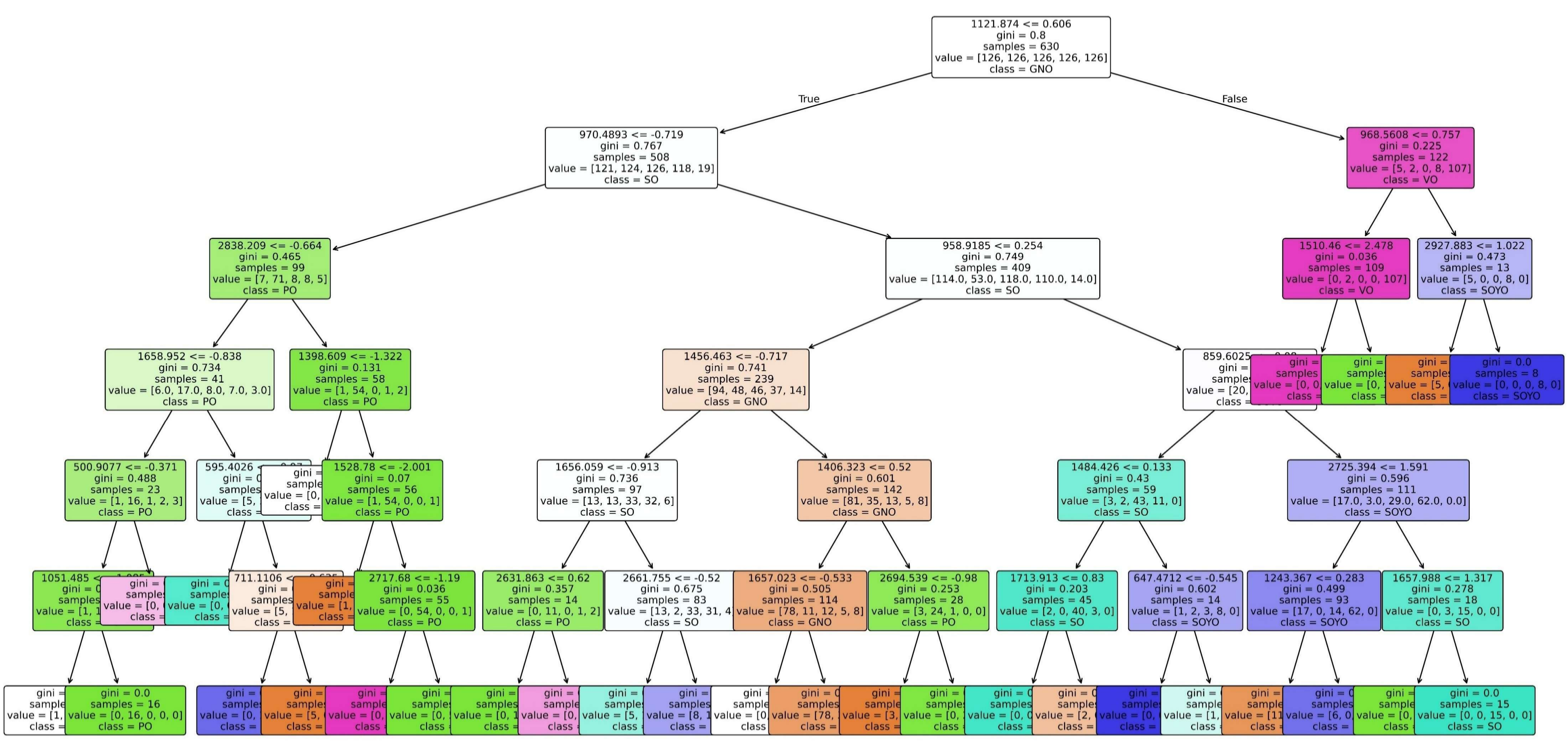

**Figure S4.** Complete graphical representation of the optimized pre-pruned Decision Tree developed using the original chips Raman dataset. The tree contains multiple decision branches and utilizes approximately 29 Raman variables, reflecting the increased classification complexity arising from food-matrix interference. The corresponding feature-importance summary is shown in **Figure 8(a)**.

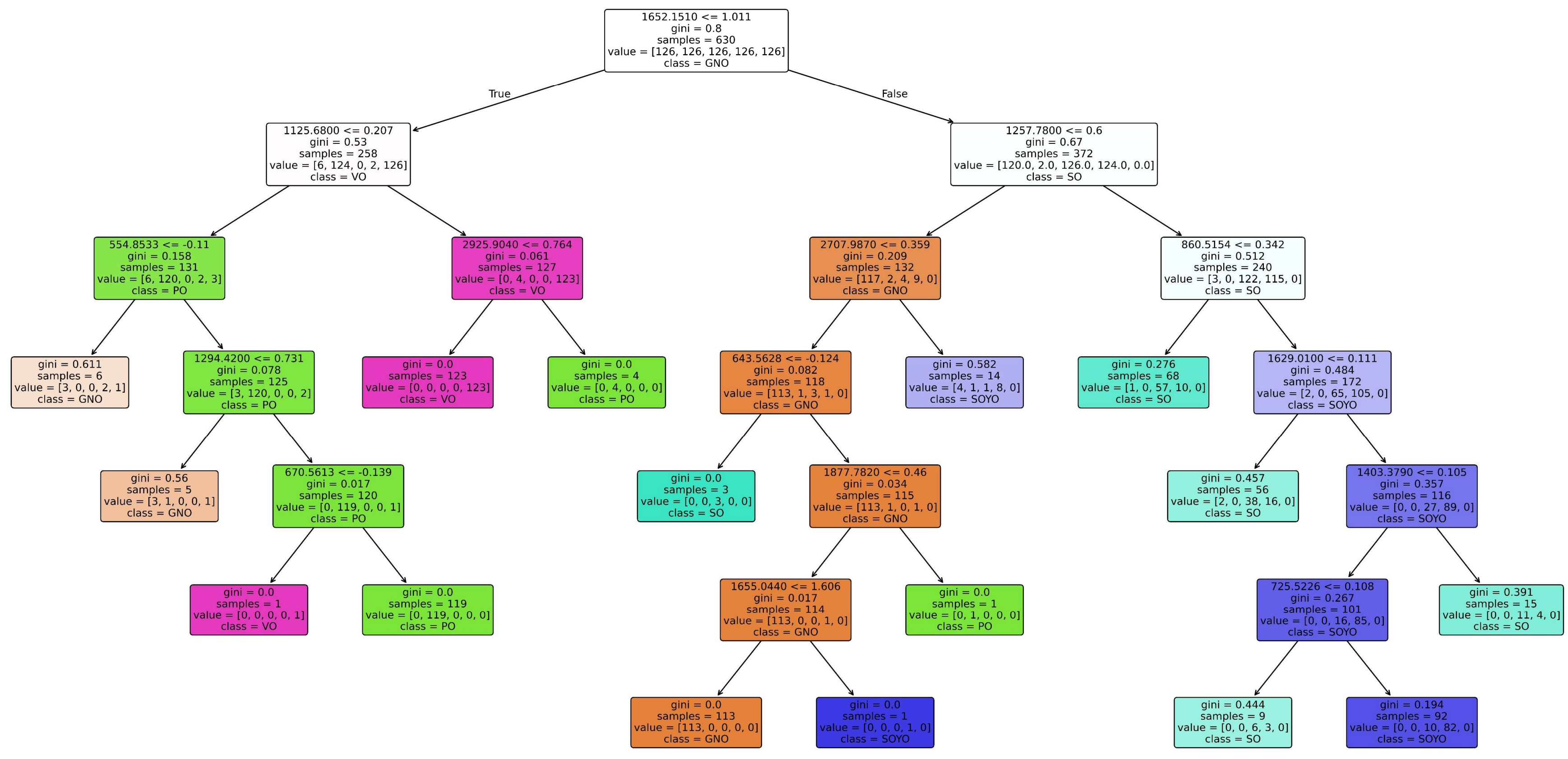

**Figure S5.** Complete graphical representation of the optimized pre-pruned Decision Tree developed using the paper-subtracted chips dataset. Following NNLS-based removal of paper contributions, the classifier becomes more compact and relies on approximately 15 important Raman variables. The corresponding feature-importance profile is presented in **Figure 8(b)**.

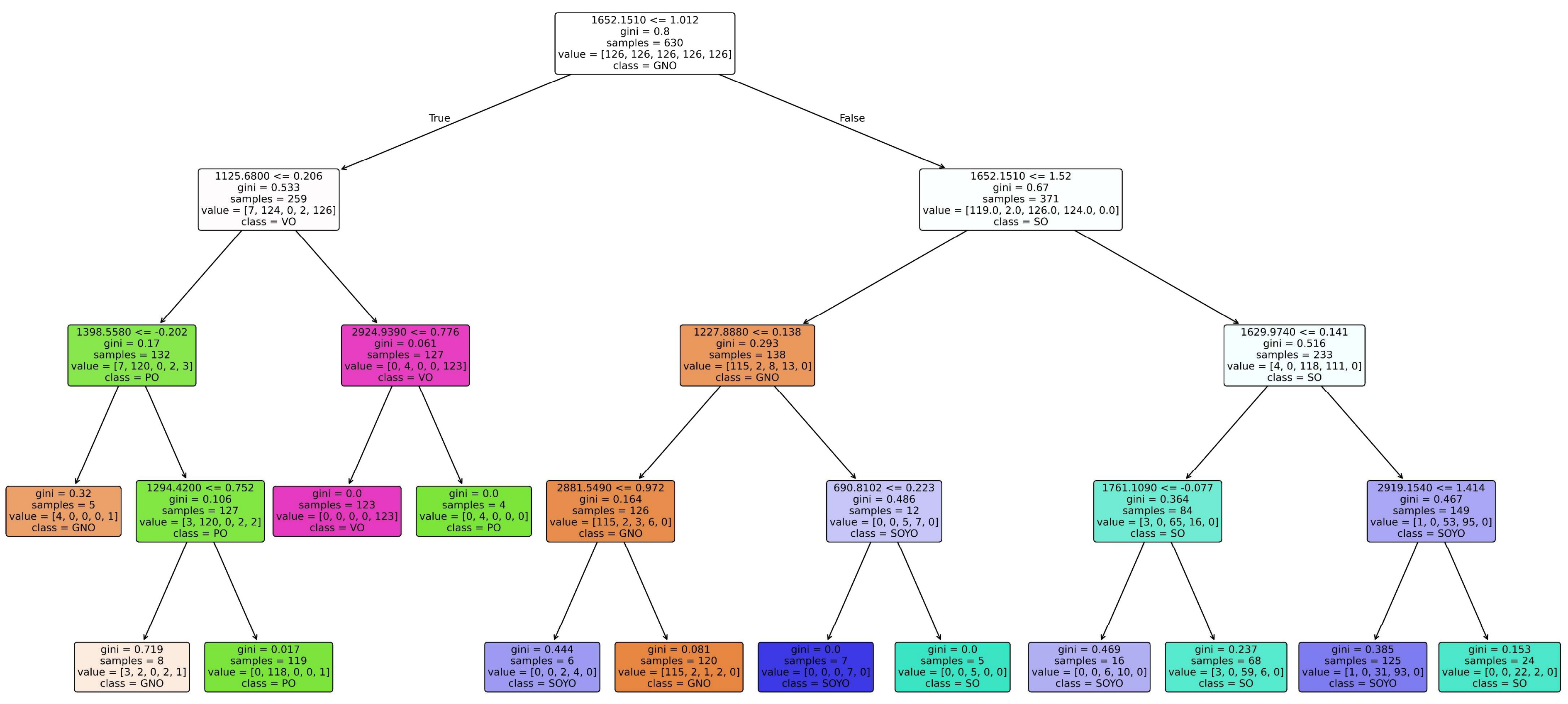

**Figure S6.** Complete graphical representation of the optimized pre-pruned Decision Tree developed using the paper- and potato-subtracted chips dataset. The reduced tree complexity and lower number of important variables (approximately 11) illustrate the effect of physics-informed NNLS preprocessing in improving spectral interpretability and concentrating classification information into fewer Raman features. The corresponding feature-importance profile is shown in **Figure 8(c).**

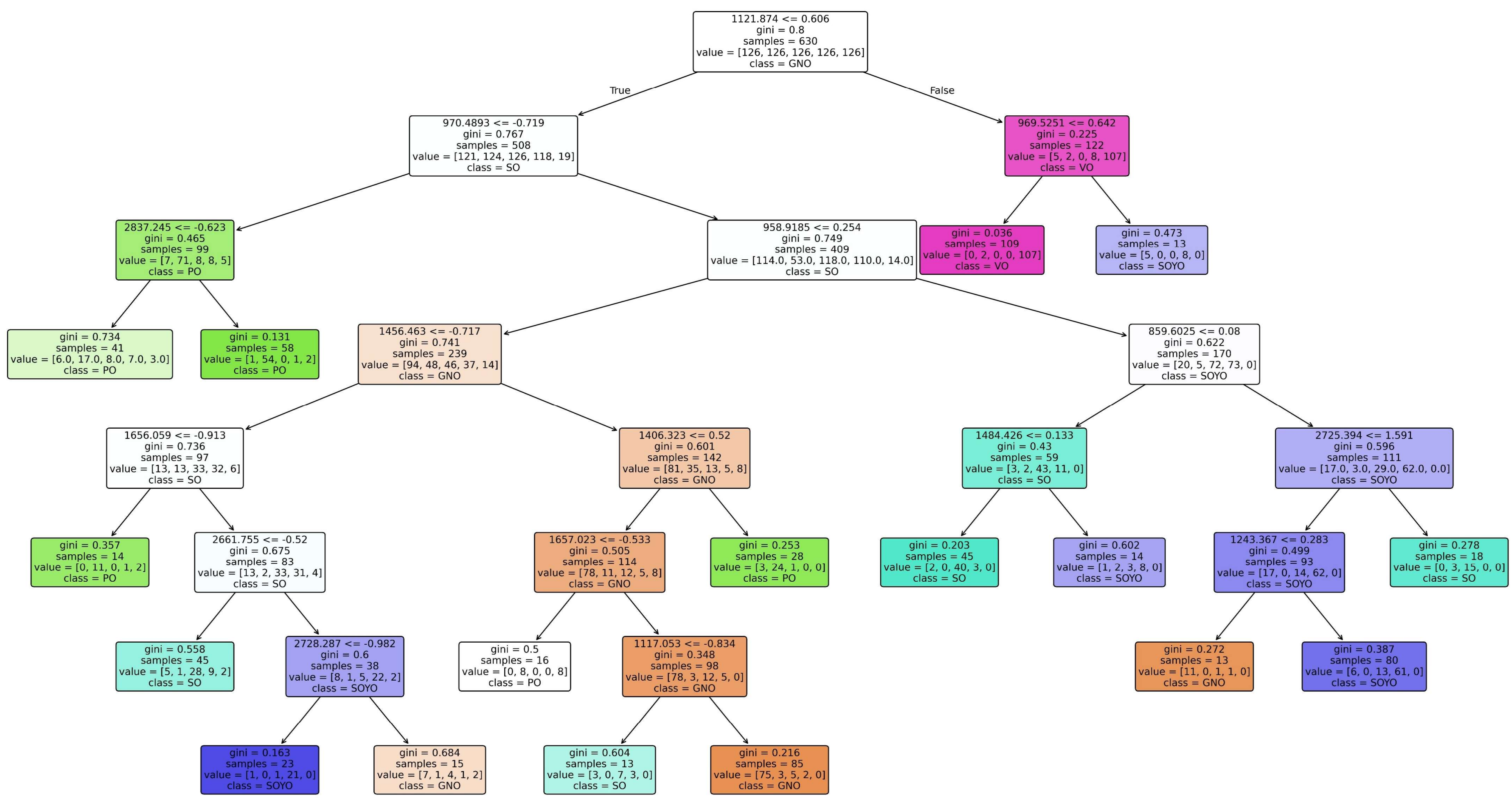

**Figure S7.** Complete graphical representation of the optimized post-pruned Decision Tree developed using the original chips Raman dataset. The classifier utilizes approximately 16 important Raman variables and provides a simplified representation of the decision boundaries used for oil classification. The corresponding feature-importance analysis is shown in **Figure 11(a)**.

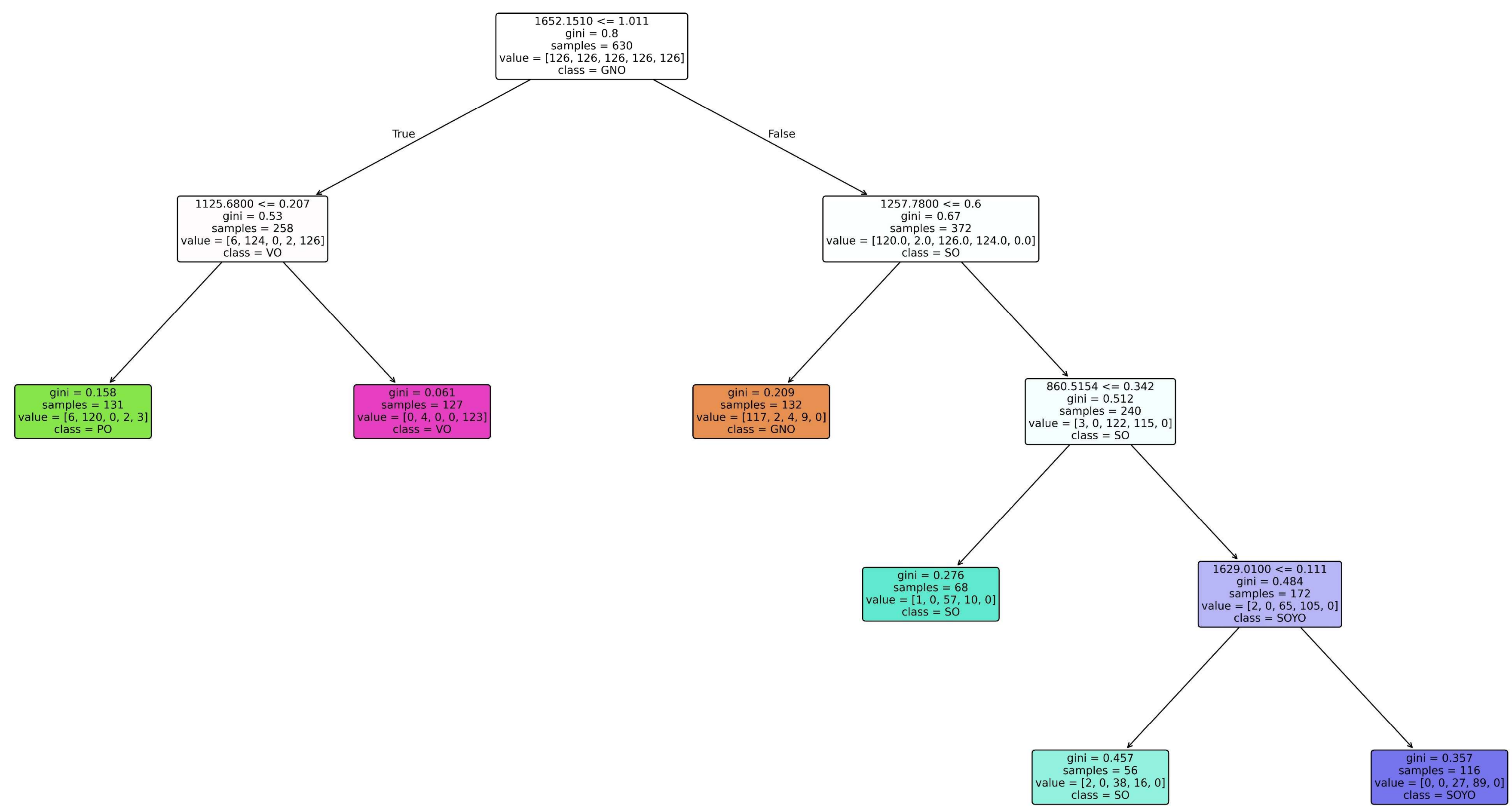

**Figure S8.** Complete graphical representation of the optimized post-pruned Decision Tree developed using the paper-subtracted chips dataset. Following NNLS-based removal of paper contributions, the classifier requires only five dominant Raman variables while achieving the highest classification accuracy among the chips' datasets. The associated feature-importance profile is presented in **Figure 11(b)**.

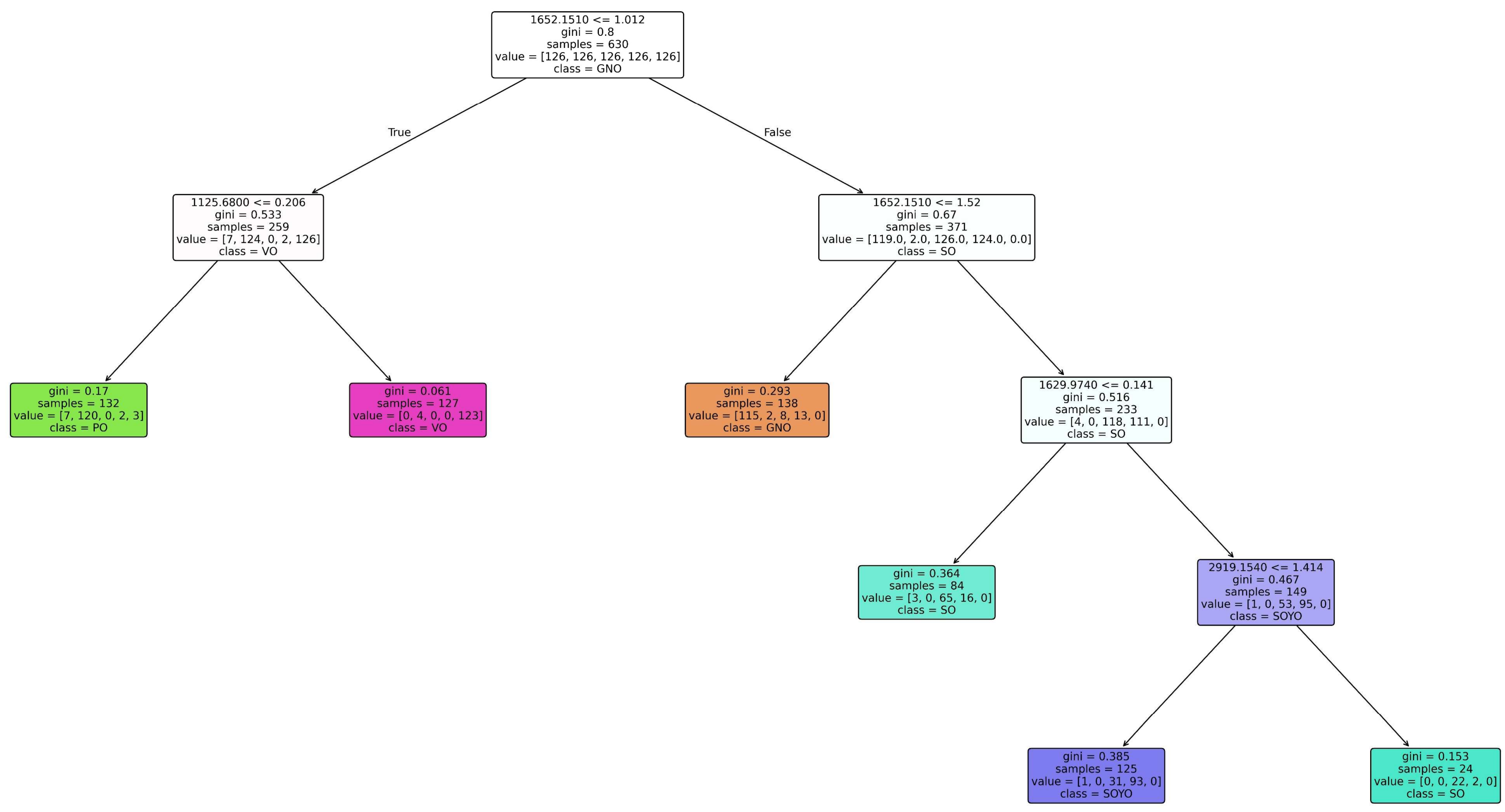

**Figure S9.** Complete graphical representation of the optimized post-pruned Decision Tree developed using the paper- and potato-subtracted chips dataset. After removal of both matrix contributions, the classifier relies on only four dominant Raman variables, demonstrating a substantial reduction in model complexity relative to the original chips' dataset. The corresponding feature-importance profile is shown in **Figure 11(c)**.